\documentclass[sigconf]{acmart}
\AtBeginDocument{%
  }
\usepackage{algorithm}
\usepackage{algpseudocode}  
\usepackage{enumerate}
\usepackage{enumitem}
\usepackage{multirow}
\usepackage{multicol}

\usepackage{tikz,pgfplots}
\usetikzlibrary{arrows.meta,positioning,calc}
\pgfplotsset{compat=1.18}
\setcopyright{acmlicensed}
\copyrightyear{2018}
\acmYear{2018}
\acmDOI{XXXXXXX.XXXXXXX}
\acmConference[Conference acronym 'XX]{Make sure to enter the correct
  conference title from your rights confirmation email}{June 03--05,
  2018}{Woodstock, NY}

\acmISBN{978-1-4503-XXXX-X/2018/06}

\begin{document}

\title{Curriculum-Guided Reinforcement Learning for Efficient Multi-Hop Retrieval-Augmented Generation}

\author{Yuelyu Ji}
\orcid{0000-0001-6389-5823}  
\affiliation{%
  \institution{University of Pittsburgh}
  \city{Pittsburgh}
  \state{PA}
  \country{USA}
}
\email{yuj49@pitt.edu}

\author{Rui Meng}
\orcid{0000-0001-5583-4924}
\affiliation{%
  \institution{Google Cloud AI Research}
  \city{Sunnyvale}
  \state{CA}
  \country{USA}
}

\author{Zhuochun Li}
\affiliation{%
  \institution{University of Pittsburgh}
    \city{Pittsburgh}
  \state{PA}
  \country{USA}
}
\author{Daqing He}
\orcid{0000-0002-4645-8696}
\affiliation{%
  \institution{University of Pittsburgh}
  \city{Pittsburgh}
  \state{PA}
  \country{USA}
}


\begin{abstract}
Retrieval-augmented generation (RAG) grounds large language models (LLMs) in up-to-date external evidence, yet existing multi-hop RAG pipelines still issue redundant sub-queries, explore too shallowly, or wander through over-long search chains. We introduce \textbf{EVO-RAG}, a curriculum-guided reinforcement-learning framework that \emph{evolves} a query-rewriting agent from broad early-stage exploration to concise late-stage refinement. EVO-RAG couples a seven-factor, step-level reward vector—covering relevance, redundancy, efficiency, and answer correctness—with a time-varying scheduler that re-weights these signals as the episode unfolds. The agent is trained with Direct Preference Optimization over a multi-head reward model, enabling it to learn when to search, back-track, answer, or refuse. Across four multi-hop QA benchmarks (HotpotQA, 2WikiMultiHopQA, MuSiQue, Bamboogle), EVO-RAG boosts Exact Match by up to \textbf{4.6 points} over strong RAG baselines while trimming average retrieval depth by \textbf{15 \%}. Ablations confirm the complementary roles of curriculum staging and dynamic reward scheduling. EVO-RAG thus offers a general recipe for building reliable, cost-effective multi-hop RAG systems.


\end{abstract}



\received{20 February 2007}
\received[revised]{12 March 2009}
\received[accepted]{5 June 2009}

\maketitle
\begin{figure}[t]
  \centering
  \includegraphics[width=0.45\textwidth]{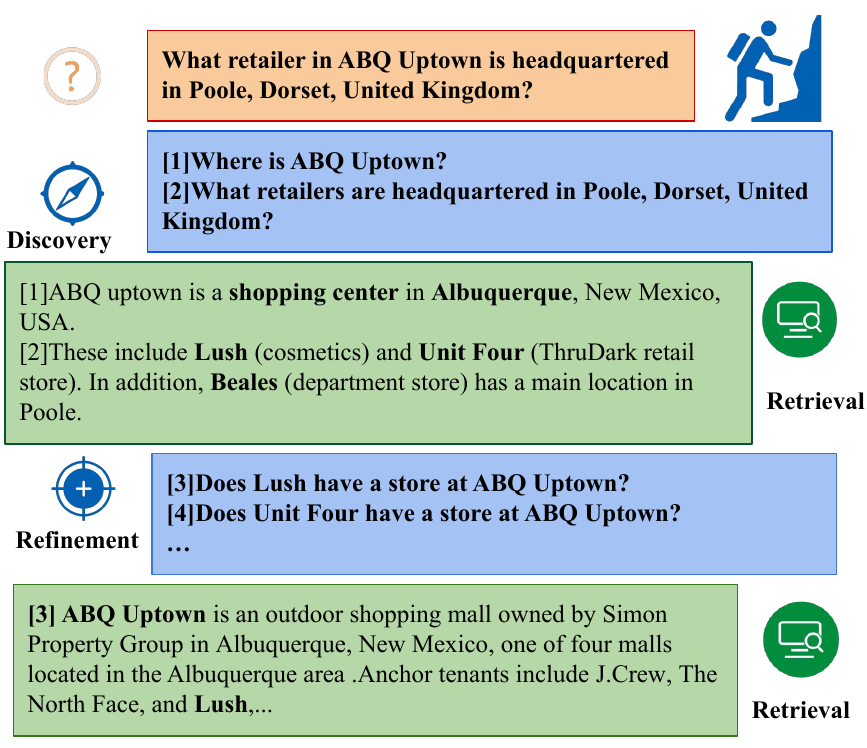}
  \caption{Illustration of EVO-RAG's two-stage curriculum. In the initial \emph{Discovery} stage, the agent broadly explores multiple retrieval pathways to identify potentially relevant documents. Subsequently, in the \emph{Refinement} stage, the agent fine-tunes queries to produce concise, evidence-backed answers. }
  \label{fig:reasoning}
\end{figure}

\section{Introduction}
Large Language Models (LLMs) have significantly advanced natural language processing, demonstrating impressive capabilities across various tasks, including question answering \cite{NEURIPS2020_1457c0d6}, conversational agents \cite{ouyang2022training}, and content generation \cite{raffel2020exploring}. Despite their strengths, LLMs frequently produce hallucinations and inaccuracies, primarily due to their reliance on static, pre-trained knowledge, which can become outdated and lacks context-specific details \cite{lewis2020retrieval}. Retrieval-Augmented Generation (RAG) addresses these limitations by dynamically incorporating external knowledge sources into the generation process, enhancing factual grounding and reducing hallucination risks \cite{lewis2020retrieval}.A multi-hop QA instance typically requires a system—or more precisely, a decision-making agent—to issue a chain of sub-queries, retrieve intermediate evidence, and reason over these clues step by step.

A typical RAG system comprises several interrelated modules, including query rewriting, document retrieval, document filtering, and answer generation \cite{gao2024smartrag,chen2024benchmarking}.In the multi-hop setting, these decisions are inter-dependent across timesteps, making a monolithic SFT objective ill-suited for global optimisation. However, while early RAG systems trained each module in isolation, recent studies have begun to explore \emph{end-to-end} objectives that jointly fine-tune the retriever and generator~\cite{gao2024smartrag,chen2024benchmarking}. Nevertheless, the prevailing supervised objectives remain static and phase-agnostic, causing goal misalignment in multi-hop reasoning. This kind of modular isolation often results in goal misalignment, negatively impacting overall coherence and accuracy \cite{xiong2025rag}. Recent approaches have introduced reinforcement learning (RL) techniques to unify these components, emphasizing cooperative optimization towards end-task objectives. Notably, MMOA-RAG employs multi-agent cooperative RL for \textbf{end-to-end optimisation} of retriever, reranker and generator, aligning them to a single performance metric~\cite{chen2025improving}.
 Similarly, frameworks like RAG-RL leverage curriculum learning and specialized policy optimization algorithms to handle complex multi-hop reasoning tasks. However, they still rely on episode-level rewards and static weighting, which are ill-equipped to drive a query-rewriting policy that must first \emph{discover} diverse evidence and later \emph{refine} its reasoning path \cite{huang2025rag}.

Despite these advances, existing RL-based RAG approaches remain \emph{phase-agnostic}. Episode-level optimisation focuses on the final answer while ignoring intermediate retrieval quality, query redundancy and computational efficiency~\cite{song2025r1}. Furthermore, their static reward schedules fail to adapt as the agent’s uncertainty shrinks, providing little guidance for the critical transition from broad exploration to conservative refinement. Consequently,  such agents tend to either under-explore or generate redundant sub-queries degrading effectiveness \cite{liu2025roserag, sun2025rearter}.

Motivated by these unresolved issues, we propose EVO-RAG, a novel evolving RAG framework that  operates in two explicit
phases—\emph{Discovery} followed by \emph{Refinement}—as illustrated in
Figure~\ref{fig:reasoning}.
During the \emph{Discovery} stage, EVO-RAG emphasizes exploratory behaviors, prioritizing retrieval breadth and query diversity to comprehensively identify relevant evidence. Subsequently, in the \emph{Refinement} stage, EVO-RAG shifts its focus to efficiency and accuracy, fine-tuning the retrieval and generation processes towards producing concise, evidence-backed answers. EVO-RAG addresses the shortcomings of prior approaches through two key innovations. First, we introduce a \textit{multi-objective reward mechanism} that explicitly captures intermediate action quality, penalizing redundant queries, backtracking inefficiencies, and poor retrieval actions while simultaneously rewarding informative and novel retrieval outcomes. Second, we implement a \textit{dynamic, time-based reward scheduler} that adjusts reward component weights throughout each retrieval-generation episode. This dynamic reward adjustment allows EVO-RAG to shift progressively from broad initial discovery to precise late-stage refinement, emphasizing efficiency and final-answer accuracy.

We conduct extensive evaluations across four widely recognized multi-hop question answering benchmarks—HotpotQA~\cite{yang2018hotpotqa}, 2WikiMultiHopQA~\cite{xanh2020_2wikimultihop}, MuSiQue~\cite{trivedi2022musique}, and Bamboogle~\cite{press2022measuring}. Our results demonstrate that EVO-RAG achieves substantial improvements over state-of-the-art methods, simultaneously enhancing accuracy and significantly reducing retrieval redundancy and computational overhead. Comprehensive ablation studies further underscore the importance of individual reward components and the effectiveness of dynamic scheduling in refining retrieval strategies.

Overall, EVO-RAG provides a robust methodological advancement for multi-stage, multi-objective optimization in RAG systems, offering a versatile and scalable framework applicable to broader NLP tasks. This research not only bridges existing gaps in RAG but also lays a solid foundation for future exploration into more sophisticated adaptive reward structures and retrieval-generation synergies.

\begin{figure*}[t]
  \centering
  \includegraphics[width=\textwidth]{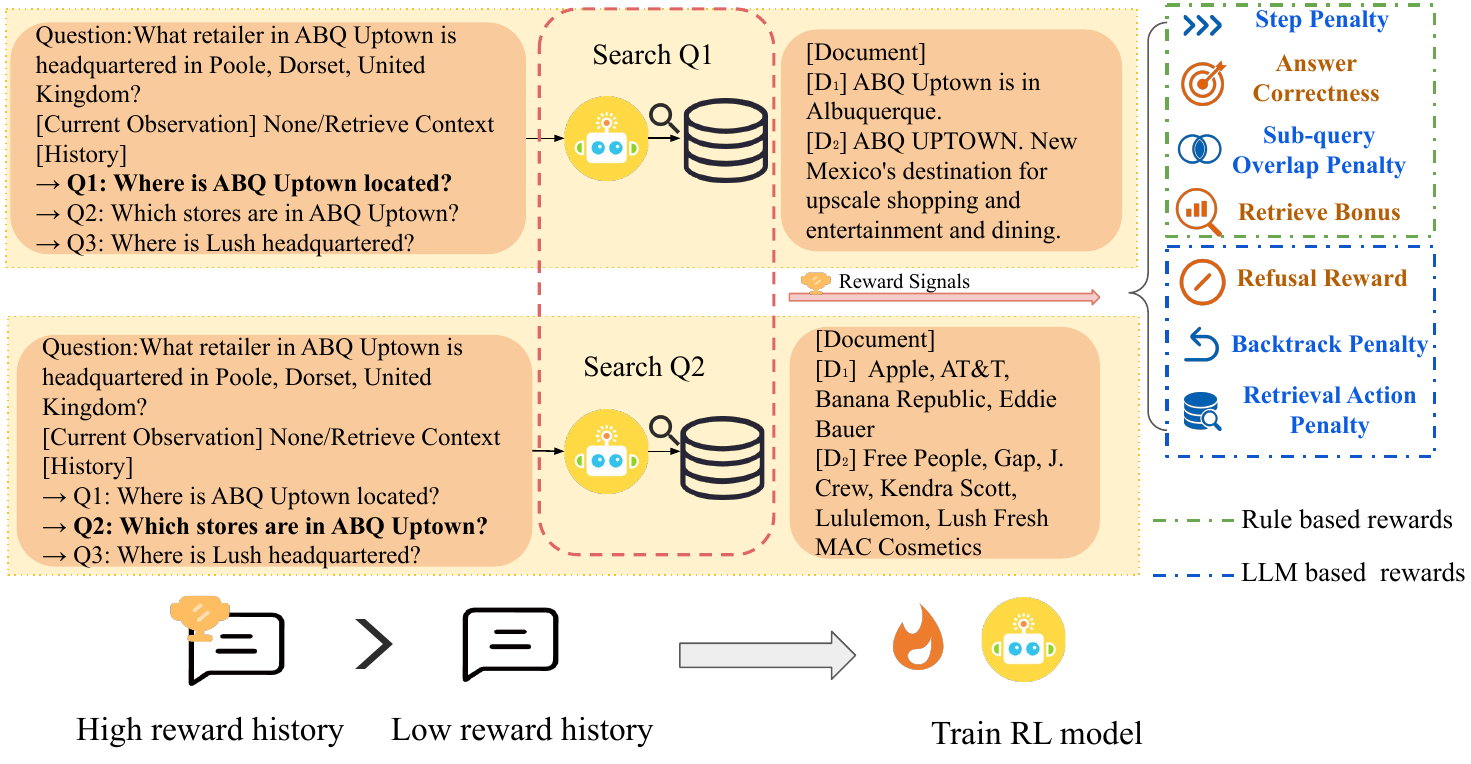}
  \caption{%
The query rewriting agent (top) interacts with the environment through four high-level actions and observes retrieved evidence at each step. 
Seven reward signals (middle) provide dense step-level feedback based on relevance, redundancy, efficiency, and final correctness.
These signals are used to train a multi-head preference model and update the agent policy via Direct Preference Optimization (DPO, bottom).
A two-stage curriculum shifts weight from early \emph{exploration} to late \emph{refinement}.
}
  \label{fig:framework}
\end{figure*}

\section{Related Work}

\paragraph{RAG Paradigms}

The Retrieval Augmented Generation (RAG) framework integrates external knowledge retrieval with language model generation, significantly enhancing the factual accuracy and reducing hallucinations inherent in standalone large language models (LLMs) \cite{lewis2020retrieval}. The RAG benchmarking study systematically analyzed various retriever-generator combinations, highlighting key trade-offs between retrieval precision and generative coherence \cite{chen2024benchmarking}. SMARTRAG further extends this approach by jointly training retrieval and generation components with end-to-end environmental feedback, demonstrating improved synergy between these modules \cite{gao2024smartrag}. Additionally, RAG-Gym offers a versatile toolkit enabling fine-grained process supervision for systematically optimizing RAG-based reasoning agents \cite{xiong2025rag,ding2024enhance,deng2024composerx,liu2024graphsnapshot,liu2024llmeasyquant,liu2024mt2st,zhao2025llm}. Existing joint-training efforts still rely on static losses and do not adapt the retrieval strategy as the LLM’s information need evolves within a dialogue or a reasoning chain, leaving room for phase-aware optimisation such as ours. 

\paragraph{Query Rewriting and Multi-Hop Retrieval}
Effective query rewriting is critical for \emph{multi-hop retrieval}, where the system must gather several pieces of evidence in sequence (e.g.\ first retrieving ``Who discovered penicillin?’’, then ``What element did Alexander Fleming help identify?’’).  
Formally, each hop $t$ issues a sub-query $q_t$ conditioned on previously retrieved passages $\{d_{<t}\}$, so errors propagate downstream if $q_t$ is poorly formed. ChainRAG iteratively completes queries by incorporating missing entities across retrieval hops, mitigating common pitfalls of incomplete searches \cite{zhu2025mitigating}. MaFeRw refines query formulations through multi-faceted feedback, including retrieval scores and generation quality, optimizing queries via proximal policy optimization (PPO) \cite{wang2024maferw}. The \mbox{IRCoT} framework~\cite{trivedi2022interleaving} prompts an LLM to generate a chain-of-thought, inserts a retrieval call whenever the thought contains a masked entity, and feeds the fetched passage back into the context before the next reasoning token.Speculative RAG enhances efficiency by simultaneously drafting multiple candidate queries with specialized models and subsequently validating these with a generalist model, substantially reducing retrieval latency \cite{wang2024speculative}. SiReRAG integrates semantic similarity and entity coherence metrics to fortify retrieval against noisy queries \cite{zhang2024sirerag}. Moreover, InstructRAG leverages explicit model-generated rationales to stabilize and enhance query rewriting \cite{wei2024instructrag,xiang2024neural,su2022mixed,zhao2025optimizedpathplanninglogistics,zhang2025automated,yang2025data,jin2024scam,jin2025scalability,li2024advances}. These methods mainly rely on heuristic rewrite rules or static reward weights; they rarely decide \emph{when to stop exploring} versus \emph{when to refine}, a gap our two-stage scheduler explicitly addresses.

\paragraph{Reinforcement Learning \& Reward Modeling in RAG}
Recent advancements integrate reinforcement learning (RL) methods into RAG to optimize retrieval-generation pipelines more holistically. MMOA-RAG adopts multi-agent cooperative RL, enabling distinct modules to collaboratively optimize towards unified performance objectives \cite{chen2025improving}. RAG-RL employs curriculum learning coupled with Group Relative Policy Optimization (GRPO) for robust multi-hop reasoning, effectively balancing exploration and exploitation \cite{huang2025rag}. Preference-based frameworks, such as PRAISE, utilize direct preference optimization methods, leveraging human-aligned feedback signals to enhance conversational question answering \cite{kaiser2025preference}. RoseRAG applies margin-aware preference optimization, significantly improving small-scale LLM robustness against noisy retrieval results \cite{liu2025roserag}. Similarly, R1-Searcher incentivizes search capabilities within LLMs using tailored RL objectives, systematically enhancing the retrieval effectiveness \cite{song2025r1}. RAG-Reward integrates RL with human feedback and automated reward models to explicitly reduce hallucinations and enhance factual correctness \cite{zhang2025rag}. Further, RbFT introduces robust fine-tuning techniques targeting retrieval defects, improving resilience against noisy retrieval scenarios \cite{tu2025rbft}.Prior RL-based RAG frameworks optimise episode-level rewards and adopt fixed weightings, which hampers their ability to penalise redundant hops or promote late-stage precision—challenges we tackle with dynamic, step-level rewards.

\paragraph{Foundation Models and Agent Architectures}
Foundation models such as LLaMA provide open, efficient bases widely adopted within RAG research, enabling scalable exploration of retrieval-augmented architectures \cite{touvron2023llama}. The GPT-4 technical report outlines robust evaluation paradigms essential for assessing large-scale model capabilities across diverse tasks \cite{achiam2023gpt}. Architectures inspired by the ReAct framework, which integrates reasoning and action in a unified LLM-based agent, provide interactive and adaptive retrieval agents, supporting sequential retrieval-generation interactions \cite{yao2023react}.

\paragraph{Retriever–Generator–Verification Frameworks}
Retriever Generator Verification (RGV) architectures enhance factual coherence by systematically generating and verifying candidate evidence documents in a structured pipeline, complementing conventional modular RAG systems \cite{SUN2025104147}. Trustworthy alignment frameworks further augment RAG models by explicitly modeling and optimizing factual reliability through specialized RL techniques \cite{zhang2024trustworthy}.

\section{Methodology}
\label{sec:method}

\subsection{Framework Overview}
\label{subsec:overview}

This section details the learning framework that forms the foundation of \textsc{EVO-RAG}, which is summarized in Figure~\ref{fig:framework}. The agent iteratively issues sub-queries, receives document feedback, and chooses between actions such as continuing, backtracking, answering, or refusing. A seven-part reward structure evaluates each step to guide fine-tuning via preference optimization.

\subsection{Step-Level Reward}
\label{sec:rewards}
In our formulation, the query–rewriting policy is treated as a 
\emph{reinforcement-learning model}. At each hop $t$, the agent observes
a state $s_t = (q_{<t}, D_{<t}, A_{<t})$, where $q_{<t}$ is the list of
previous sub-queries, $D_{<t}$ the corresponding retrieved documents,
and $A_{<t}$ any intermediate answers and rationales. The agent then selects an action $a_t$ from a discrete space. 
$$\mathcal{A} = \{\texttt{SEARCH}, \texttt{BACKTRACK}, \texttt{ANSWER}, \texttt{REFUSE}\}$$

In multi-hop RAG, the quality of each retrieval and query rewriting step directly impacts final answer accuracy. To guide the model’s behaviour at each step, EVO-RAG defines a
seven-dimensional \emph{step-level reward vector}
$\mathbf{r}_t = (r_t^{(1)}, \dots, r_t^{(7)})$.
We now introduce and motivate two core components.

\paragraph{Retrieval Bonus ($r_{\text{ret}}$).}  
At each step $t$, the agent selects an action $a_t \in \mathcal{A}$ and retrieves a document set $D_t$ if $a_t = \texttt{SEARCH}$.  
Let $\mathcal{D}^*$ denote the gold answer-supporting documents.
To reward successful retrieval, we define:
\[
r_{\text{ret}}(s_t,a_t)=
  \begin{cases}
    +1 & \text{if } a_t = \texttt{SEARCH} \land D_t \cap \mathcal{D}^* \neq \varnothing, \\[2pt]
    -1 & \text{if } a_t = \texttt{SEARCH} \land D_t \cap \mathcal{D}^* = \varnothing, \\[2pt]
     0 & \text{otherwise}.
  \end{cases}
\]
This reward evaluates the quality of each \texttt{SEARCH} action based on whether it retrieves any relevant document.
Although the step index $t$ does not directly appear in the reward formula, the agent accumulates $r_{\text{ret}}$ \emph{at every step}.
Hence, discovering relevant evidence earlier allows the agent to gather more positive signal across the episode, incentivizing early and effective retrieval.

\textbf{Note.} $r_{\text{ret}}$ is \emph{time–agnostic}: it only checks
whether the current \texttt{SEARCH} hits a gold document; the earlier the
hit, the more positive reward the agent accumulates.

\paragraph{Sub-query Overlap Penalty ($r_{\text{dup}}$).}  
A prevalent inefficiency in multi-hop RAG is that the model issues
nearly identical sub-queries, wasting retrieval calls without bringing
new evidence.  
Let $\mathbf q_t = f_{\text{enc}}(q_t)$ be the embedding of the current
sub-query and $\mathbf q_j$ the embedding of a previous sub-query
($j<t$). We use the sentence-transformers/all-MiniLM-L6-v2 \cite{wang2020minilm} to embedding the queries. We penalise the maximum cosine similarity:
\[
r_{\text{dup}}(s_t,a_t)= -\max_{j<t}\cos\!\bigl(\mathbf q_t,\mathbf q_j\bigr).
\]
The larger the overlap with any earlier query, the stronger the
negative reward, encouraging diversity while still permitting similarity
when genuinely useful.

\paragraph{Back-track Penalty ($r_{\text{bt}}$).}  
Back-tracking decisions are made by the LLM based policy itself:
whenever the model selects the high-level action \texttt{BACKTRACK},
we apply a fixed penalty
\[
r_{\text{bt}}(s_t,a_t)= -\mathbf 1\!\bigl[a_t=\texttt{BACKTRACK}\bigr],
\]
where $\mathbf 1[\cdot]$ is the indicator function.
The negative reward discourages gratuitous back-tracking, yet allows
the policy to pay the cost when it predicts that returning to a
previous state will yield higher cumulative reward downstream.

\paragraph{Refusal Reward ($r_{\text{ref}}$).}  
In practice, \textit{unanswerable} is not an oracle flag but the
judgement of an external LLM verifier (ChatGPT-4o) that inspects the
current evidence set $D_{\le t}$ and returns \texttt{enough\_evidence}
$\in\{\texttt{yes},\texttt{no}\}$.  Thus the policy is rewarded for \emph{truthfully refusing} when the verifier deems the evidence insufficient, rather than for skipping
retrieval to save cost.
\[
r_{\text{ref}}(s_t,a_t)=
\begin{cases}
 +1 & a_t=\texttt{REFUSE}\land\text{unanswerable},\\[2pt]
 -1 & a_t=\texttt{REFUSE}\land\text{answerable},\\[2pt]
  0 & \text{otherwise}.
\end{cases}
\]

\paragraph{Step Cost ($r_{\text{step}}$).}

Excessively long reasoning chains harm computational efficiency and user experience. We introduce a uniform negative reward per step to discourage unnecessarily long query sequences:
\[
r_{\text{step}}(s_t,a_t) = -1.
\]
Although the raw penalty is $-1$, its effective contribution is modulated by a dynamic weight $w_{\text{step}}(t)$, which linearly increases from $0.02$ to $0.10$ as the reasoning progresses (see Table~\ref{tab:weights}). This ensures the penalty grows as the episode lengthens, without imposing a large deduction per step early on.  
We also experimented with alternative raw values ($-0.5$, $-2$), but $-1$ yielded the best balance between answer accuracy (–0.3 EM drop vs.\ $-0.5$) and average step count (–1.1 steps vs.\ $-0.5$), and was thus adopted.

\paragraph{Answer Correctness ($r_{\text{ans}}$).}  
Ultimately, the goal of retrieval-augmented QA is to produce accurate answers. To ensure the entire retrieval-generation process aligns with this objective, we apply a correctness reward only at the termination step ($T$):
\[
r_{\text{ans}}(s_T,a_T)=\frac{1}{2}\left[\text{EM}(A_T,A^*)+\text{F1}(A_T,A^*)\right],
\]
where $A_T$ is the model’s final prediction and $A^*$ is the ground-truth answer. $\text{EM}$ denotes exact match accuracy, while $\text{F1}$ measures token-level overlap. Although $\text{F1}$ partially includes EM cases, their average balances strict correctness with partial informativeness, encouraging the model to produce responses that are both accurate and semantically complete.

\paragraph{Retrieval Action Penalty ($r_{\text{act}}$).}
This term prices the \emph{cost} of issuing additional
\texttt{SEARCH} actions and grows as the episode advances.  It is
independent of correctness, which is already captured by
$r_{\text{ret}}$.

\[
r_{\text{act}}(s_t,a_t)=
\begin{cases}
\,0                                 & a_t=\texttt{SEARCH}\,\land\,p(t)<0.3,\\[4pt]
\,-\mathbf 1[r_{\text{dup}}<0]      & a_t=\texttt{SEARCH}\,\land\,p(t)\ge 0.3,\\[4pt]
\,0                                 & \text{otherwise}.
\end{cases}
\]

\noindent
where $p(t)=\tfrac{t}{T}$ and $r_{\text{dup}}$ is the overlap penalty.
Thus \emph{early} searches are free, but \emph{late} and
\emph{redundant} ones pay a penalty whose weight
$\lambda(t)$ increases linearly from $0.4$ to $1.2$ (Table~\ref{tab:weights}).

\paragraph{Aggregated Reward.}
All step-level rewards are linearly combined to yield the total signal:
\[
R_t = \sum_{i} w_i(t) \cdot r_i(s_t, a_t),
\]


where $r_i(\cdot)$ denotes the $i$-th reward component and $w_i(t)$ is a time-dependent weight. These weights are annealed across the episode (see Table~\ref{tab:weights}), enabling the policy to emphasize different objectives at different reasoning phases. For instance, $w_{\text{step}}(t)$ increases from $0.02$ to $0.10$, making the step penalty stronger in later stages. We empirically tuned each $w_i$ to balance effectiveness and efficiency; ablation studies in Section~\ref{sec:ablation} highlight the contribution of each reward term.

\subsection{Multi‑Head Preference Model}
\label{sec:rm}

Following the process-supervision recipe of RAG-Gym\,\cite{xiong2025rag},
we learn a preference function $f_\phi$ that ranks sibling actions
based on their cumulative reward rather than regressing raw scalars.
For each visited state $s_t$, we construct pairwise preferences
$(x^+, x^-)$ from two trajectory branches ending in different actions,
where $x^+$ denotes the branch with higher total reward.
To ensure training stability, we only retain comparisons where the
absolute reward difference exceeds a threshold:
\[
|R(x^+) - R(x^-)| \geq \Delta,
\quad\text{with } \Delta = 0.3.
\]
Here, $R(x) = \sum_{t} w_t \cdot r_t(s_t, a_t)$ denotes the
cumulative weighted return of trajectory $x$, computed using the
time-dependent reward weights described in Section~\ref{sec:scheduler}.

A frozen LLM encoder(eg. LLaMA-3.1-8B) embeds each trajectory prefix, and
\textbf{seven parallel linear heads}
$\{f^{(k)}_\phi\}_{k=1}^7$ (one per reward) output scalar scores.
We minimize the averaged pairwise logistic loss:
\[
\mathcal{L}_{\text{RM}} =
  -\tfrac{1}{7} \sum_{k=1}^7
  \log \sigma\bigl(f^{(k)}_\phi(x^+) - f^{(k)}_\phi(x^-)\bigr),
\]
where $\sigma$ is the sigmoid function.
This loss reduces to Direct Preference Optimization (DPO) with
temperature~$1$ if all heads are tied.

Compared to a single-head reward model trained on the total reward,
our multi-head variant achieved lower validation loss
(e.g., 0.62 vs.\ 0.71) and more stable policy updates.

\begin{figure}[htbp]
  \centering
  \includegraphics[width=0.45\textwidth]{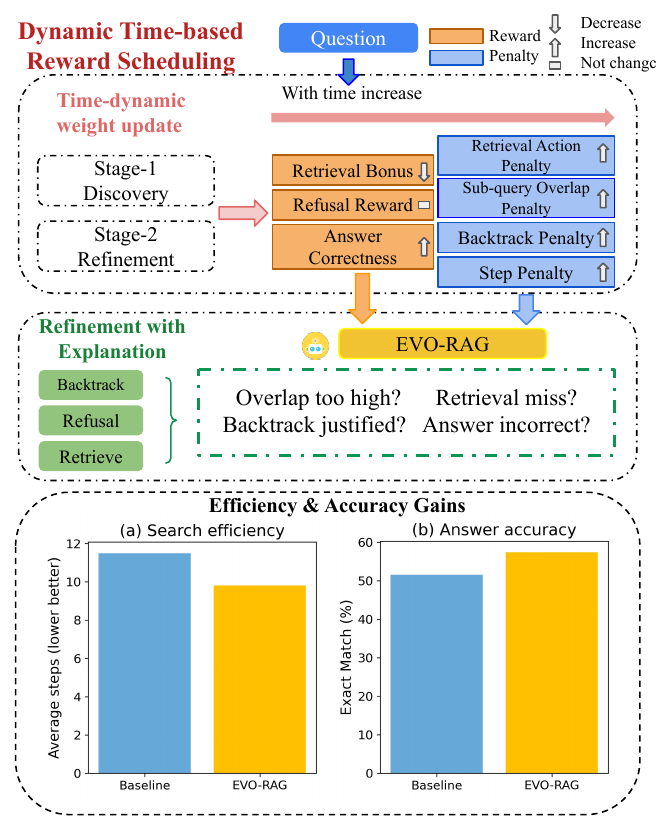}
  \caption{Reward weights for EVO-RAG training. Stage 1 and Stage 2 represent the Discovery and Refinement phases, respectively. Arrows indicate weight trends.}
  \label{fig:evo-rag-overview}
\end{figure}

\subsection{Policy Optimisation}
\label{sec:policy}

Policy learning alternates three steps:

\begin{enumerate}[label=(\roman*)]
  \item \emph{Roll‑out}: generate $N$ trajectories under the current curriculum.
  \item \emph{Preference extraction}: at each state select one positive/negative
        action pair with margin $\delta=0.2$.
  \item \emph{Update}: minimise
        \[
          \mathcal{L}_{\text{DPO}}
          =-\log\sigma\!\bigl(\beta\,[\log\pi_\theta(x^+)-\log\pi_\theta(x^-)]\bigr),
          \quad\beta=0.1.
        \]
\end{enumerate}

We fine‑tune LoRA adapters with effective batch 64
(8‑fold accumulation) and learning rate $5\times10^{-6}$,
alternating one reward‑model epoch with one policy epoch.
Training stops if HotpotQA dev EM stalls for two cycles.

\subsection{Two‑Stage Curriculum}
\label{sec:twostage}
\paragraph{Notation for weight vectors.}
We denote the stage-dependent weight vector as
$\boldsymbol{w}(t)=(\beta,\lambda,\gamma,\delta,\rho,\eta,\kappa)$,
where each symbol is the coefficient for the corresponding reward
defined in Section~\ref{sec:rewards}:
$\beta$—Retrieval Bonus,
$\lambda$—Retrieval Action Penalty,
$\gamma$—Sub-query Overlap Penalty,
$\delta$—Back-track Penalty,
$\rho$—Refusal Reward,
$\eta$—Step Penalty,
$\kappa$—Answer Correctness.
Our full training pipeline is visualized in Figure~\ref{fig:evo-rag-overview}. It illustrates the curriculum design and time-dependent reward weighting that underpins our two-stage process.
Specifically, training is split into two clearly defined stages:
a \textbf{Discovery} stage followed by a \textbf{Refinement} stage.
During the \emph{Discovery} stage, we emphasize retrieval-oriented reward components $(\beta,\lambda)$ to encourage the agent to explore diverse queries thoroughly. Efficiency-related rewards $(\eta,\gamma,\kappa)$ remain relatively small, allowing the agent to freely explore and accumulate a broad set of potentially relevant documents.

After achieving stable convergence during the \emph{Discovery} stage, we transition into the \emph{Refinement} stage. In this phase, we reload the best-performing checkpoint from the previous stage and shift the reward priorities: we significantly amplify efficiency-oriented and accuracy-oriented rewards $(\kappa,\gamma,\eta)$, while gradually reducing the importance of retrieval-oriented rewards $(\beta,\lambda)$. This encourages the agent to refine its query-generation strategy, focusing on concise, targeted queries that reliably support accurate and evidence-backed answers.
Table~\ref{tab:weights} lists the early, mid and late weights that
anchor the schedule; the next subsection explains how they are interpolated
\emph{within} each episode.

\subsection{Time‑Based Scheduler}
\label{sec:scheduler}

Inside every episode the reward weights evolve with the
progress ratio $p(t)=t/T_{\max}$, where $T_{\max}=20$:

\[
w_k(t)
  =(1-p(t))\,w_k^{\text{early}}+p(t)\,w_k^{\text{late}}.
\]
\begin{table}[t]
  \centering
  \small
  \renewcommand{\arraystretch}{1.1}
  \setlength{\tabcolsep}{4pt}
  \caption{Reward weights for EVO-RAG training. "Start" to "Mid" columns represent the interpolation range during Stage 1 (Discovery), and "Mid" to "End" represent Stage 2 (Refinement). Arrows ($\nearrow$, $\searrow$) indicate increasing or decreasing weight trends.}
  \label{tab:weights}
  \begin{tabular}{@{}lccc|ccc@{}}
    \toprule
    \multirow{2}{*}{Reward Component} 
    & \multicolumn{3}{c|}{Stage 1: Discovery} 
    & \multicolumn{3}{c}{Stage 2: Refinement} \\ 
    \cmidrule(lr){2-4}\cmidrule(lr){5-7}
    & Start & Mid & Trend & Mid & End & Trend \\
    \midrule
    Retrieval Bonus ($\beta$)              & 2.0  & 1.0  & $\searrow$ & 1.0  & 0.5  & $\searrow$ \\
    Retrieval Action Penalty ($\lambda$)   & 1.5  & 0.8  & $\searrow$ & 0.8  & 0.4  & $\searrow$ \\
    Subquery Overlap Penalty ($\gamma$)    & 0.1  & 0.5  & $\nearrow$ & 0.5  & 1.2  & $\nearrow$ \\
    Backtrack Penalty ($\delta$)           & 0.3  & 0.5  & $\nearrow$ & 0.5  & 1.0  & $\nearrow$ \\
    Refusal Reward ($\rho$)                & 0.5  & 0.5  & --         & 0.5  & 0.5  & --         \\
    Step Penalty ($\eta$)                  & 0.02 & 0.05 & $\nearrow$ & 0.05 & 0.10 & $\nearrow$ \\
    Answer Correctness ($\kappa$)          & 0.05 & 0.10 & $\nearrow$ & 0.10 & 1.00 & $\nearrow$ \\
    \bottomrule
  \end{tabular}
\end{table}
Retrieval‑focused weights $(\beta,\lambda)$ monotonically decrease,
whereas efficiency‑focused weights $(\gamma,\eta,\kappa)$ increase;
the refusal weight $\rho$ stays constant.
This “gearbox” provides step‑level guidance that is missing from
a static two‑stage switch.

\subsection{Overall Training Loop}
\label{sec:loop}

Algorithm~\ref{alg:evo} summarizes the complete training loop of EVO-RAG, which alternates between reward-model rollouts and policy optimization across two curriculum stages. The same training logic is re-used for both the \textit{Discovery} and \textit{Refinement} phases. With these components combined, EVO‑RAG delivers the dense,
curriculum‑guided feedback necessary for efficient multi‑hop retrieval.

\begin{algorithm}[t]
  \caption{EVO‑RAG training loop}
  \label{alg:evo}
  \footnotesize
  \begin{algorithmic}[1]
    \State Initialise $\pi_\theta$ \Comment{e.g., from supervised warm start}
    \For{stage $\in\{\text{Discovery},\text{Refinement}\}$}
      \For{$m=1$ \textbf{to} $M$ episodes}
        \State Reset environment; $t\leftarrow0$
        \While{$t<T_{\max}$ \textbf{and} not terminal}
          \State Observe $s_t$; sample $a_t\sim\pi_\theta(\cdot|s_t)$
          \State Execute $a_t$; receive $r_t$; update weights $w_k(t)$ \Comment{dynamic weight update}
          \State $t\leftarrow t+1$
        \EndWhile
        \State Store trajectory and sibling preferences
      \EndFor
      \State Fine‑tune $\pi_\theta$ for $E$ epochs with DPO \Comment{based on preference pairs}
    \EndFor
  \end{algorithmic}
\end{algorithm}

\noindent

\begin{table}[h] 
\centering 
\caption{Dataset Statistics}
\label{tab:dataset-stats}
\begin{tabular}{lcc} 
\toprule Dataset & Eval Size & Task Style \\
\midrule HotpotQA& 7,404  & Factoid multi-hop QA \\
2WikiMultiHopQA & 12,575 & Structured multi-hop QA \\
MuSiQue & 2417 & Compositional multi-hop QA \\
Bamboogle & 125  & Adversarial multi-hop QA \\
\bottomrule 
\end{tabular} 
\end{table}

\begin{table*}[ht]
  \centering
  \setlength{\tabcolsep}{5pt}
  \caption{Comparison of RAG Methods on Multi-hop QA Datasets. Metrics are Exact Match (EM) / F1, averaged over 3 seeds.}
  \label{tab:main-results}
  \begin{tabular}{lccccc}
    \toprule
    \textbf{Method (best setting in paper)} &
    \textbf{Backbone} &
    \textbf{HotpotQA} &
    \textbf{2WikiMultiHopQA} &
    \textbf{MuSiQue} &
    \textbf{Bamboogle} \\
    \midrule
    RAG-Gym (ReSearch + PRM) \cite{xiong2025rag} & LLaMA-3.1-8B
        & 44.1 / 56.8
        & 50.2 / 57.9
        & 48 / 60
        & 51.2 / 63.1 \\

    IRCoT\cite{trivedi2022interleaving} + Flan-T5-XXL      & Flan-T5-XXL
  & 45.0 / 56.2
  & 45.4 / 56.8
  & 19.9 / 24.9
  & 44.0 / 55.0 \\
    \midrule
    \textbf{EVO-RAG} & DeepSeek-8B
        & 57.8 / 71.4
        & 52.6 / 66.4
        & 51.8 / 63.7
        & 45.3 / 58.2 \\

    \textbf{EVO-RAG} & LLaMA-3.1-8B
        & 57.4 / 71.2
        & 53.0 / 66.9
        & \textbf{52.5 / 64.4}
        & 45.7 / 58.6 \\

    \textbf{EVO-RAG} & Qwen-3-8B
        & \textbf{57.6 / 71.5}
        & \textbf{53.2 / 67.1}
        & 52.2 / 64.0
        & \textbf{46.0 / 59.0} \\
    \bottomrule
  \end{tabular}
  \label{tab:main-results-complete}
\end{table*}


\begin{table}[h]
\centering
\caption{HotpotQA results under different reward schedules.}
\label{tab:reward-schedule-compact}
\begin{tabular}{llcc}
\toprule
\textbf{Backbone} & \textbf{Strategy} & \textbf{EM} & \textbf{F1} \\
\midrule
\multirow{3}{*}{DeepSeek-8B}
    & No Reward & 52.6\% & 66.2\% \\
    & Two-stage & 55.0\% & 68.7\% \\
    & \textbf{Time-dynamic} & \textbf{56.8\%} & \textbf{70.5\%} \\
\midrule
\multirow{3}{*}{LLaMA-3.1-8B}
    & No Reward & 52.9\% & 66.6\% \\
    & Two-stage &  \textbf{57.4\%} &\textbf{71.2\%}  \\
    & \textbf{Time-dynamic} &55.6\% &69.4\%  \\
\midrule
\multirow{3}{*}{Qwen3-8B}
    & No Reward & 53.1\% & 66.7\% \\
    & Two-stage & 55.9\% & 69.5\% \\
    & \textbf{Time-dynamic} & \textbf{57.6\%} & \textbf{71.5\%} \\
\bottomrule
\end{tabular}
\end{table}

\begin{table}[h]
\centering \caption{Single-Reward Ablation Results} 
\label{tab:single-reward}
\begin{tabular}{lcc} 
\toprule Single Reward Type & Eval Accuracy (\%) & Eval Loss \\
\midrule Backtrack Reward & \textbf{70.31} & 0.913 \\
Refusal Reward & 60.58 & 1.018 \\
Retrieve Reward & 55.24 & 1.089 \\
Step Penalty & 54.17 & 1.184 \\
Subquery Overlap Penalty & 54.35 & 1.015 \\
\bottomrule
\end{tabular}
\end{table}

\begin{figure}[h]
  \centering
  \includegraphics[width=0.9\linewidth]{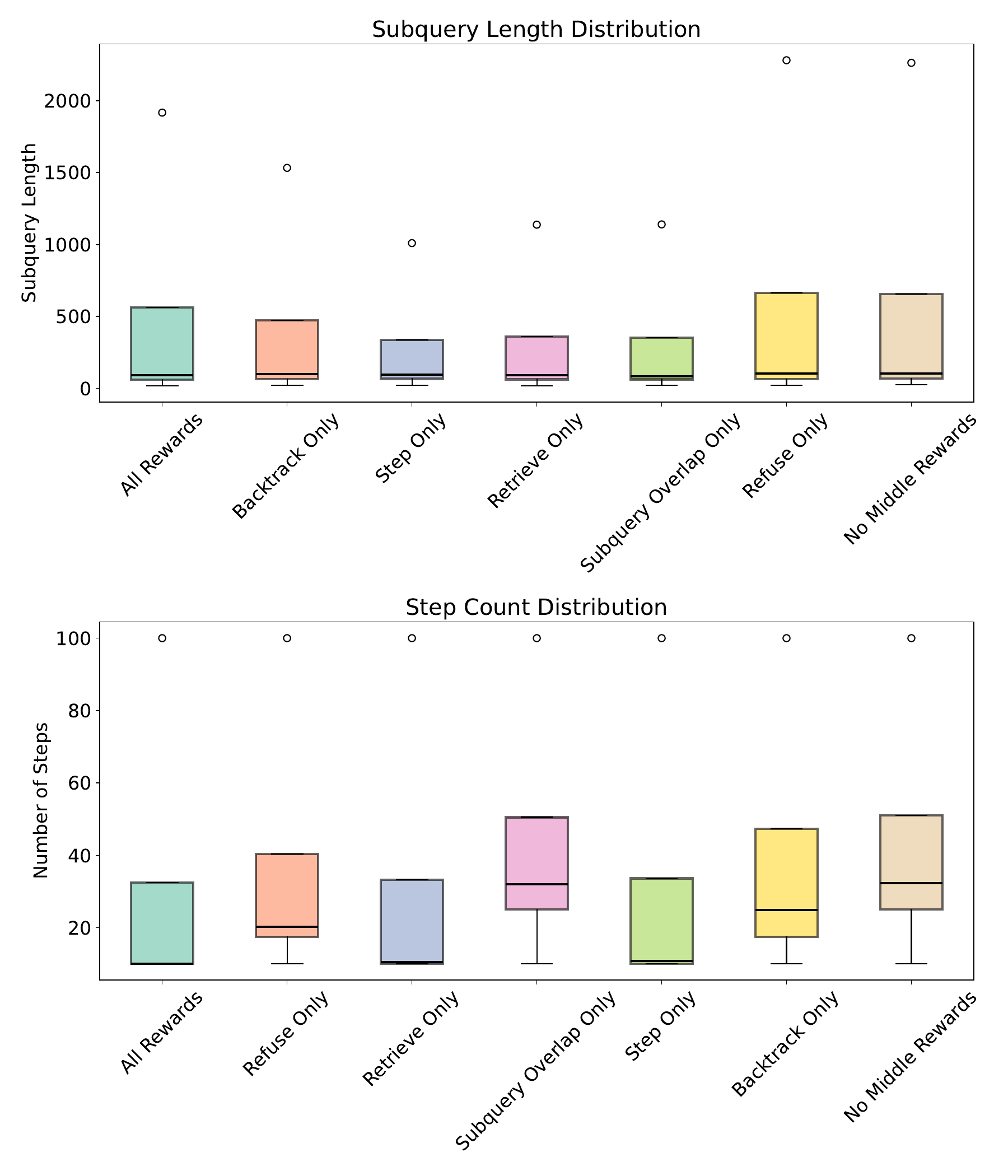}
  \caption{Sub-query length (top) and step count (bottom) distributions under various reward configurations.}
  \label{fig:reward-boxplot}
\end{figure}
\begin{table}[h]
\centering
\caption{Impact of different reward \emph{combinations} on HotpotQA using Qwen3-8B. Metrics: Exact Match (EM) / F1; Avg.\ Steps indicates average retrieval length.}
\label{tab:reward-combo}
\small
\begin{tabular}{lccc}
\toprule
\textbf{Reward Combination} & \textbf{EM (\%)} & \textbf{F1 (\%)} & \textbf{Avg.\ Steps} \\
\midrule
Baseline (No Reward)          & 53.1 & 66.7 &  8.2 \\
Best-2 (Backtrack + AnsCorr)   & 56.2 & 70.0 & 11.3 \\
Best-3 (+Overlap)              & 56.9 & 70.6 & 10.1 \\
Exploration-Heavy              & 55.0 & 69.1 & 13.4 \\
Efficiency-Heavy               & 55.4 & 68.8 &  9.0 \\
\textbf{Full (All Rewards)}    & \textbf{57.6} & \textbf{71.5} & 10.4 \\
\bottomrule
\end{tabular}
\end{table}

\section{Experiments and Results}
\label{sec:experiments}

We systematically evaluate EVO-RAG on four prominent multi-hop QA benchmarks: HotpotQA, 2WikiMultiHopQA, MuSiQue, and Bamboogle. Our evaluation specifically targets the following three research questions:
\begin{enumerate}[label=\textbf{RQ\arabic*:}, leftmargin=12pt]
    \item \emph{Does the two-stage, dynamic curriculum improve retrieval and answer quality over strong RAG baselines?}
    \item \emph{How sensitive is the agent performance to individual reward components?}
    \item \emph{What benefits does our time-based scheduler offer over static or simple stage-switching methods?}
\end{enumerate}

\subsection{Datasets and Setup}
We evaluate EVO-RAG on four multi-hop QA benchmarks (Table~\ref{tab:dataset-stats}). All models are trained using 1,000 queries sampled from HotpotQA. Evaluation is conducted on official validation sets. Answer generation is evaluated using Exact Match (EM) and F1 scores. We intentionally keep training confined to HotpotQA to test cross-dataset generalisation; see Section \ref{sec:case} for qualitative evidence.

\paragraph{LLM Backbone and Retriever.}
We use Meta-Llama-3.1-8B-Instruct \cite{touvron2023llama}, Qwen3-8B \cite{yang2025qwen3}, Deepseek-R1-distill-llama3-8B \cite{guo2025deepseek} as the agent backbone, paired with RRF-BGE \cite{chen2024bge} retriever (fusion of BM25 \cite{robertson2009probabilistic} and BGE embeddings).

\paragraph{Training Setup.}
Rollout generation is conducted in two phases (explore and refine) with dynamic reward interpolation (§\ref{sec:method}). We adopt LoRA adapters for parameter-efficient fine-tuning, and use DPO loss for preference optimization.

\subsection{Main Results}
Table~\ref{tab:main-results-complete} comprehensively compares EVO-RAG with two state-of-the-art multi-hop QA baselines, RAG-Gym~\cite{xiong2025rag} and ReARTeR~\cite{sun2025rearter}, across multiple challenging datasets. EVO-RAG consistently achieves superior performance, notably improving Exact Match (EM) and F1 scores across all datasets. Specifically, on the widely recognized benchmark HotpotQA, Compared with the \emph{interleaved} baseline IRCoT, EVO-RAG–Qwen3-8B gains +12.6 EM and +15.3 F1 on HotpotQA, and +7.8 EM / +10.3 F1 on 2WikiMultiHopQA, confirming that curriculum-based RL is substantially more effective than fixed heuristic alternation. The substantial enhancements are similarly observed on the 2WikiMultiHopQA dataset (+1.7 EM, +1.9 F1), highlighting the robustness and generalization capabilities of EVO-RAG. Additionally, EVO-RAG attains notable gains on the more complex and adversarial MuSiQue and Bamboogle datasets, underscoring its efficacy in challenging retrieval environments.

\paragraph{Baseline (No Reward)} This clearly illustrates the performance ceiling when relying solely on standard supervised training without step-level guidance or explicit penalties for inefficient retrieval behavior.
It relies solely on the original RAG formulation without any explicit optimization towards query efficiency, retrieval accuracy, or answer correctness beyond standard supervised training. This setup highlights the limitations of a purely retrieval-agnostic approach. EVO-RAG significantly outperforms previous best methods, achieving +2.0\% EM and +1.8\% F1 improvements over ReARTeR~\cite{sun2025rearter} on HotpotQA.
\begin{table*}[h]
  \centering
  \small
  \caption{Detailed trajectories (\# steps, retrieved evidence, and actions) under different reward scheduling strategies. ``Dup.'' = near‑duplicate, $\Diamond$ indicates the timestep where the reward model preferred the chosen action. Correct answers are \textbf{bold}; wrong ones are marked in \textcolor{red}{red}.}
  \label{tab:case-detail}
  \begin{tabular}{@{}p{2.9cm}p{4.6cm}p{4.6cm}p{4.6cm}@{}}
    \toprule
    & \textbf{Baseline (No Reward)} & \textbf{Two-stage (Fixed)} & \textbf{Time-dynamic (EVO-RAG)} \\
    \midrule[0.15em]
    \multicolumn{4}{@{}l}{\textbf{Q1: “In which year was the monarch who issued the 1925 Birthday Honours born?”}} \\ \midrule
    Steps & 1 & 2 & 2 \\
    $q_1$ & Birthday Honours 1925 monarch birth & same as Baseline & Who was monarch of the UK in 1925? \\
    Doc$_1$ & \texttt{…George V – 1867–1936 – Birthday Honours… (unreliable wiki list)} & \texttt{same noisy list} & \texttt{George V – King of the UK 1910–1936; born 3 Jun 1865} \\
    $q_2$ & – & When was George V born? (Dup.$\Diamond$) & When was George V born? \\
    Doc$_2$ & – & \texttt{George V – born 1865} & \texttt{George V – born 1865} \\
    Action & \textcolor{red}{1867} (hallucinated) & \textbf{1865} & \textbf{1865} \\
    \midrule[0.15em]
    \multicolumn{4}{@{}l}{\textbf{Q2: “Which U.S. state contains the launch site of Mars Pathfinder?”}} \\ \midrule
    Steps & 1 & 6 & 2 \\
    $q_1$ & Mars Pathfinder launch site state & Mars Pathfinder launch pad location & Launch site of Mars Pathfinder? \\
    Doc$_1$ & \texttt{…launched from Cape Canaveral LC‑17…} & \texttt{Cape Canaveral Air Force Station…} & \texttt{launched from Cape Canaveral SLC‑17A…} \\
    $q_{2..5}$ & – & multiple paraphrases about “launch pad” (Dup.) & Which state is Cape Canaveral in? \\
    Doc$_k$ & – & repeated retrievals & \texttt{Cape Canaveral – city in Florida, U.S.} \\
    Action & \textcolor{red}{California} & \textbf{Florida} & \textbf{Florida} \\
    \midrule[0.15em]
    \multicolumn{4}{@{}l}{\textbf{Q3: “Where was the 2021 Hugo Award ceremony hosted?” (labelled unanswerable)}} \\ \midrule
    Steps & 1 & 13 & 4 \\
    $q_1$ & 2021 Hugo Award host city & same & same \\
    $q_{2..n}$ & – & repeated variants (Dup.) & variants but stopped timely \\
    Evidence & no reliable doc & repetitive sci-fi convention lists & no relevant doc after multiple tries \\
    Action & \textcolor{red}{London} (no citation) & \textcolor{red}{Dublin} & \textit{REFUSE} (\checkmark) \\
    \bottomrule
  \end{tabular}
\end{table*}

\subsection{Impact of Dynamic Reward Scheduling}

To evaluate the impact of the dynamic reward scheduling strategy employed in EVO-RAG, we conduct an extensive comparison against a baseline (No Reward) and a fixed two-stage scheduling (Table~\ref{tab:reward-schedule-compact}). Across all three evaluated backbones (DeepSeek-8B, LLaMA-3.1-8B, and Qwen3-8B), the introduction of dynamic, time-based scheduling generally leads to better or comparable accuracy relative to the fixed two-stage scheduling. Notably, for DeepSeek-8B and Qwen3-8B, the dynamic scheduling consistently enhances accuracy by approximately +1.8 EM and F1 on average. Interestingly, the LLaMA-3.1-8B backbone exhibits slightly lower accuracy under dynamic scheduling compared to the fixed two-stage scheduling. This may suggest specific interactions between this model's inherent learning dynamics and the adaptive reward schedule, which could benefit from further investigation.

Table~\ref{tab:reward-combo} further clarifies these findings by demonstrating the average number of retrieval steps. The full dynamic reward setup ("All Rewards") attains the optimal balance between retrieval efficiency and accuracy, achieving the highest EM (57.6\%) and F1 (71.5\%) while maintaining a reasonable average step length of 10.4. Conversely, a purely exploration-heavy reward setup results in longer chains (13.4 steps) with slightly lower accuracy (55.0 EM), emphasizing the importance of balancing exploration with precision in retrieval.

\subsection{Single-Reward Training Study}
\label{sec:ablation}

We train agents using only a single reward type to investigate individual signal strength (Table~\ref{tab:single-reward}).
Observation: The backtrack signal alone achieves a high internal evaluation accuracy (70.3\%), defined as the agent's accuracy in selecting preferable retrieval actions according to the learned preference model, indicating that search flexibility is critical for robust multi-hop retrieval.

Figure~\ref{fig:reward-boxplot} clearly illustrates that the full reward configuration reduces both the median retrieval step count and variance compared to other configurations, indicating more consistent and efficient retrieval behavior across queries. To better understand how individual reward signals interact, we evaluate several carefully selected reward \emph{combinations} on HotpotQA (Table~\ref{tab:reward-combo}).  
Three observations emerge:

\paragraph{(i) Core signals are complementary.}  
Combining \textit{Backtrack} with \textit{AnswerCorrectness} (Best-2) already brings a +3.1 EM gain over the no-reward baseline, confirming that flexible back-tracking must be coupled with an answer-level objective to yield benefits.

\paragraph{(ii) Overlap penalties improve efficiency without hurting accuracy.}  
Adding \textit{Sub-query Overlap Penalty} (Best-3) further improves EM/F1 and reduces the average reasoning depth from 11.3 to 10.1 steps, indicating that discouraging redundant queries directly translates into more concise retrieval chains.

\paragraph{(iii) Full reward set offers the best trade-off.}  
While an \textit{Exploration-Heavy} setup maximizes recall(Retrieval Bonus ($r_{ret}$) + Sub-query Overlap Penalty ($r_{dup}$)), it causes longer trajectories and slightly lower accuracy.  
Conversely, an \textit{Efficiency-Heavy} mix shortens trajectories but sacrifices EM (Backtrack Reward ($r_{\text{bt}}$) + Step Penalty ($r_{\text{step}}$)).  
Our full, time-dynamic reward suite delivers the highest overall accuracy (57.6\% EM) while maintaining an efficient retrieval trajectory (10.4 steps on average). Although this is slightly higher than the shortest average step count observed (10.1 steps with Best-3), the marginal increase is justified by significant accuracy gains, indicating an optimal trade-off between exploration, accuracy, and efficiency.

\section{Qualitative Case Studies}
\label{sec:case}

To show how different reward–guidance schemes influence retrieval behaviour, we analyse three HotpotQA questions and compare three model variants: (i) a \emph{Baseline} that fires one sub‑query then answers, (ii) a \emph{Two‑stage} curriculum with fixed reward weights inside each phase, and (iii) our best model, the \emph{Time‑dynamic} curriculum implemented with \textsc{EVO‑RAG~Qwen3‑8B} (hereafter \textsc{EvoQ3}).  For every question we present the ordered sub‑queries $q_t$, the top retrieved snippet, and the final action (answer or refusal).

\subsection{Reward Guidance Recap}
\textbf{Baseline.} This variant optimises only the final answer‑correctness reward ($\kappa$).  It has no step cost ($\eta$), no refusal signal ($\rho$) and no overlap penalty ($\gamma$).  Consequently the agent has no incentive to continue searching after its first hit—regardless of evidence quality—and never refuses to answer.

\textbf{Two‑stage (fixed).} Training is split into an discovery phase where retrieval bonus ($\beta$) dominates and a refinement phase where correctness ($\kappa$) is up‑weighted.  Inside each phase the weights are constant, so the agent receives identical feedback at the first and the tenth step of the same episode.  This setting promotes early breadth but offers little pressure to stop issuing redundant queries later on.

\textbf{Time‑dynamic.} On top of the two global stages, each reward component is linearly interpolated inside every episode: $\beta$ and the retrieval‑cost penalty ($\lambda$) decay with step ratio $p\!=\!t/T_{\max}$, while $\gamma$, $\eta$ and $\kappa$ rise.  A dedicated refusal reward $\rho$ is always active.  The gradual shift lets the agent explore in the first few steps, then tightens efficiency and precision constraints as evidence accumulates.

\subsection{Step‑level Traces}
For each question we show (see Table~\ref{tab:case-detail})  (1) the ordered sub‑queries $q_t$, (2) the top retrieved passage title and a short snippet, and (3) the agent action ($\rightarrow$ \textbf{A}: answer or \textit{Refuse}).
\vspace{4pt}

\begin{enumerate}
  \item \textbf{Failure‑to‑refine in Baseline.}  With only one query and no step penalty, the agent latches onto the first noisy snippet and hallucinates (Q1/Q2).
  \item \textbf{Late duplication in Two‑stage.}  Because overlap penalty $\gamma$ is flat within a phase, redundant paraphrases persist (Q2), and the agent chases non‑existent evidence on unanswerable questions (Q3).
  \item \textbf{Balanced behaviour in Time‑dynamic.}  Decaying $\beta$ prevents endless searching, rising $\gamma$ suppresses duplicates, and $\rho$ enables safe refusal—yielding correct answers (Q1/Q2) and honest abstention (Q3) with fewer steps.
\end{enumerate}

\paragraph{Metrics Alignment.}  These traces mirror quantitative gains: Time dynamic improves EM by +1.7 over Two‑stage on HotpotQA while cutting average steps from 11.5 to 9.8 (cf.\ Table~\ref{tab:reward-combo}).

\paragraph{Observation 1: Time‑dynamic scheduler curbs duplication.} In Q1 and Q2 the overlap penalty $\gamma$ grows with the episode step, discouraging redundant paraphrases that still plague \textsc{2Stage}.  The reward model distinguishes the concise path at step~$q_2$ ($\Diamond$).

\paragraph{Observation 2: Multi‑aspect rewards foster caution.}  Q1 shows how the combination of retrieval bonus $\beta$ and correctness reward $\kappa$ motivates \textsc{Evo-RAG-Time} to \emph{verify} the monarch before emitting an answer, unlike \textsc{BackOnly}.

\paragraph{Observation 3: Refusal signal prevents hallucination.}  On unanswerable Q3 the specialised refusal reward $\rho$ leads \textsc{Evo-RAG-Time} to abstain, whereas baselines hallucinate.

These case studies underline how curriculum‑guided, multi‑facet supervision translates into qualitatively superior reasoning behaviour.

\section{Conclusion}
EVO-RAG addresses a fundamental limitation in existing RAG systems by dynamically optimizing the sequence of sub-queries during multi-hop retrieval. By integrating a seven-dimensional, dynamically weighted reward structure and employing a multi-head preference model, EVO-RAG effectively balances exploratory and exploitative retrieval behaviors. Our extensive evaluation across four diverse benchmarks demonstrates substantial gains in both accuracy (up to +4.6 points in Exact Match) and efficiency (reducing retrieval steps by approximately 15\%). Qualitative analyses reinforce these findings, highlighting EVO-RAG’s ability to reduce redundant queries, utilize strategic backtracking, and appropriately refuse to answer when evidence is insufficient. Overall, EVO-RAG showcases the potential of dynamically structured reward systems to enhance the reliability and efficiency of RAG pipelines.

\section{Limitations}
While EVO-RAG achieves notable performance gains, several limitations remain. Our evaluation relies exclusively on automatic metrics (Exact Match and F1), thus human judgment on answer quality, factuality, and utility remains unexplored. Furthermore, reward parameters and scheduling were manually tuned primarily on the HotpotQA dataset; thus, performance may vary across domains without additional tuning. The current implementation employs explicit action prompts rather than fully learned latent actions, potentially restricting flexibility and adaptability in diverse scenarios. Lastly, computational experiments are limited to a single GPU with moderate model sizes (8B parameters); scaling to larger models or extending context lengths would increase computational costs and require further optimization.

\section{Future Work}
Several avenues of future research can enhance EVO-RAG’s adaptability and generalization. Replacing manually fixed reward schedules with adaptive or meta-learned coefficients could enable the framework to autonomously optimize reward signals across different datasets and retrieval tasks. Transitioning from explicit discrete actions to latent action policies could significantly broaden EVO-RAG's applicability to more open-ended tasks such as conversational agents and interactive dialogue systems. Extending evaluations to retrieval-augmented summarization, scientific fact verification, and specialized domain searches (e.g., legal or patent retrieval) would validate the framework’s versatility across varied retrieval scenarios. Finally, integrating calibrated uncertainty estimates into the reward structure could enable efficient early stopping and escalation mechanisms, significantly enhancing real-world reliability and user trust.
\section*{GenAI Usage Disclosure}

The authors affirm that no part of the paper's text was generated entirely by generative AI tools. Large Language Models (LLMs) were used exclusively for minor grammar editing and formatting suggestions. All code, data annotations, and scientific contributions were created by the authors. The preference model analysis and reward formulation were designed and implemented without GenAI assistance.

\bibliographystyle{ACM-Reference-Format}
\bibliography{sample-base}


\begin{thebibliography}{49}


\ifx \showCODEN    \undefined \def \showCODEN     #1{\unskip}     \fi
\ifx \showISBNx    \undefined \def \showISBNx     #1{\unskip}     \fi
\ifx \showISBNxiii \undefined \def \showISBNxiii  #1{\unskip}     \fi
\ifx \showISSN     \undefined \def \showISSN      #1{\unskip}     \fi
\ifx \showLCCN     \undefined \def \showLCCN      #1{\unskip}     \fi
\ifx \shownote     \undefined \def \shownote      #1{#1}          \fi
\ifx \showarticletitle \undefined \def \showarticletitle #1{#1}   \fi
\ifx \showURL      \undefined \def \showURL       {\relax}        \fi
\providecommand\bibfield[2]{#2}
\providecommand\bibinfo[2]{#2}
\providecommand\natexlab[1]{#1}
\providecommand\showeprint[2][]{arXiv:#2}

\bibitem[Achiam et~al\mbox{.}(2023)]%
        {achiam2023gpt}
\bibfield{author}{\bibinfo{person}{Josh Achiam}, \bibinfo{person}{Steven Adler}, \bibinfo{person}{Sandhini Agarwal}, \bibinfo{person}{Lama Ahmad}, \bibinfo{person}{Ilge Akkaya}, \bibinfo{person}{Florencia~Leoni Aleman}, \bibinfo{person}{Diogo Almeida}, \bibinfo{person}{Janko Altenschmidt}, \bibinfo{person}{Sam Altman}, \bibinfo{person}{Shyamal Anadkat}, {et~al\mbox{.}}} \bibinfo{year}{2023}\natexlab{}.
\newblock \showarticletitle{Gpt-4 technical report}.
\newblock \bibinfo{journal}{\emph{arXiv preprint arXiv:2303.08774}} (\bibinfo{year}{2023}).
\newblock


\bibitem[Brown et~al\mbox{.}(2020)]%
        {NEURIPS2020_1457c0d6}
\bibfield{author}{\bibinfo{person}{Tom Brown}, \bibinfo{person}{Benjamin Mann}, \bibinfo{person}{Nick Ryder}, \bibinfo{person}{Melanie Subbiah}, \bibinfo{person}{Jared~D Kaplan}, \bibinfo{person}{Prafulla Dhariwal}, \bibinfo{person}{Arvind Neelakantan}, \bibinfo{person}{Pranav Shyam}, \bibinfo{person}{Girish Sastry}, \bibinfo{person}{Amanda Askell}, \bibinfo{person}{Sandhini Agarwal}, \bibinfo{person}{Ariel Herbert-Voss}, \bibinfo{person}{Gretchen Krueger}, \bibinfo{person}{Tom Henighan}, \bibinfo{person}{Rewon Child}, \bibinfo{person}{Aditya Ramesh}, \bibinfo{person}{Daniel Ziegler}, \bibinfo{person}{Jeffrey Wu}, \bibinfo{person}{Clemens Winter}, \bibinfo{person}{Chris Hesse}, \bibinfo{person}{Mark Chen}, \bibinfo{person}{Eric Sigler}, \bibinfo{person}{Mateusz Litwin}, \bibinfo{person}{Scott Gray}, \bibinfo{person}{Benjamin Chess}, \bibinfo{person}{Jack Clark}, \bibinfo{person}{Christopher Berner}, \bibinfo{person}{Sam McCandlish}, \bibinfo{person}{Alec Radford}, \bibinfo{person}{Ilya Sutskever}, {and}
  \bibinfo{person}{Dario Amodei}.} \bibinfo{year}{2020}\natexlab{}.
\newblock \showarticletitle{Language Models are Few-Shot Learners}. In \bibinfo{booktitle}{\emph{Advances in Neural Information Processing Systems}}, \bibfield{editor}{\bibinfo{person}{H.~Larochelle}, \bibinfo{person}{M.~Ranzato}, \bibinfo{person}{R.~Hadsell}, \bibinfo{person}{M.F. Balcan}, {and} \bibinfo{person}{H.~Lin}} (Eds.), Vol.~\bibinfo{volume}{33}. \bibinfo{publisher}{Curran Associates, Inc.}, \bibinfo{pages}{1877--1901}.
\newblock
\urldef\tempurl%
\url{https://proceedings.neurips.cc/paper_files/paper/2020/file/1457c0d6bfcb4967418bfb8ac142f64a-Paper.pdf}
\showURL{%
\tempurl}


\bibitem[Chen et~al\mbox{.}(2024a)]%
        {chen2024benchmarking}
\bibfield{author}{\bibinfo{person}{Jiawei Chen}, \bibinfo{person}{Hongyu Lin}, \bibinfo{person}{Xianpei Han}, {and} \bibinfo{person}{Le Sun}.} \bibinfo{year}{2024}\natexlab{a}.
\newblock \showarticletitle{Benchmarking large language models in retrieval-augmented generation}. In \bibinfo{booktitle}{\emph{Proceedings of the AAAI Conference on Artificial Intelligence}}, Vol.~\bibinfo{volume}{38}. \bibinfo{pages}{17754--17762}.
\newblock


\bibitem[Chen et~al\mbox{.}(2024b)]%
        {chen2024bge}
\bibfield{author}{\bibinfo{person}{Jianlv Chen}, \bibinfo{person}{Shitao Xiao}, \bibinfo{person}{Peitian Zhang}, \bibinfo{person}{Kun Luo}, \bibinfo{person}{Defu Lian}, {and} \bibinfo{person}{Zheng Liu}.} \bibinfo{year}{2024}\natexlab{b}.
\newblock \showarticletitle{Bge m3-embedding: Multi-lingual, multi-functionality, multi-granularity text embeddings through self-knowledge distillation}.
\newblock \bibinfo{journal}{\emph{arXiv preprint arXiv:2402.03216}} (\bibinfo{year}{2024}).
\newblock


\bibitem[Chen et~al\mbox{.}(2025)]%
        {chen2025improving}
\bibfield{author}{\bibinfo{person}{Yiqun Chen}, \bibinfo{person}{Lingyong Yan}, \bibinfo{person}{Weiwei Sun}, \bibinfo{person}{Xinyu Ma}, \bibinfo{person}{Yi Zhang}, \bibinfo{person}{Shuaiqiang Wang}, \bibinfo{person}{Dawei Yin}, \bibinfo{person}{Yiming Yang}, {and} \bibinfo{person}{Jiaxin Mao}.} \bibinfo{year}{2025}\natexlab{}.
\newblock \showarticletitle{Improving Retrieval-Augmented Generation through Multi-Agent Reinforcement Learning}.
\newblock \bibinfo{journal}{\emph{arXiv preprint arXiv:2501.15228}} (\bibinfo{year}{2025}).
\newblock


\bibitem[Deng et~al\mbox{.}(2024)]%
        {deng2024composerx}
\bibfield{author}{\bibinfo{person}{Qixin Deng}, \bibinfo{person}{Qikai Yang}, \bibinfo{person}{Ruibin Yuan}, \bibinfo{person}{Yipeng Huang}, \bibinfo{person}{Yi Wang}, \bibinfo{person}{Xubo Liu}, \bibinfo{person}{Zeyue Tian}, \bibinfo{person}{Jiahao Pan}, \bibinfo{person}{Ge Zhang}, \bibinfo{person}{Hanfeng Lin}, {et~al\mbox{.}}} \bibinfo{year}{2024}\natexlab{}.
\newblock \showarticletitle{ComposerX: Multi-Agent Symbolic Music Composition with LLMs}. In \bibinfo{booktitle}{\emph{The 25th International Society for Music Information Retrieval Conference}}.
\newblock


\bibitem[Ding et~al\mbox{.}(2024)]%
        {ding2024enhance}
\bibfield{author}{\bibinfo{person}{Zhicheng Ding}, \bibinfo{person}{Panfeng Li}, \bibinfo{person}{Qikai Yang}, {and} \bibinfo{person}{Siyang Li}.} \bibinfo{year}{2024}\natexlab{}.
\newblock \showarticletitle{Enhance image-to-image generation with llava-generated prompts}. In \bibinfo{booktitle}{\emph{2024 5th International Conference on Information Science, Parallel and Distributed Systems (ISPDS)}}. IEEE, \bibinfo{pages}{77--81}.
\newblock


\bibitem[Gao et~al\mbox{.}(2024)]%
        {gao2024smartrag}
\bibfield{author}{\bibinfo{person}{Jingsheng Gao}, \bibinfo{person}{Linxu Li}, \bibinfo{person}{Weiyuan Li}, \bibinfo{person}{Yuzhuo Fu}, {and} \bibinfo{person}{Bin Dai}.} \bibinfo{year}{2024}\natexlab{}.
\newblock \showarticletitle{SmartRAG: Jointly Learn RAG-Related Tasks From the Environment Feedback}.
\newblock \bibinfo{journal}{\emph{arXiv preprint arXiv:2410.18141}} (\bibinfo{year}{2024}).
\newblock


\bibitem[Guo et~al\mbox{.}(2025)]%
        {guo2025deepseek}
\bibfield{author}{\bibinfo{person}{Daya Guo}, \bibinfo{person}{Dejian Yang}, \bibinfo{person}{Haowei Zhang}, \bibinfo{person}{Junxiao Song}, \bibinfo{person}{Ruoyu Zhang}, \bibinfo{person}{Runxin Xu}, \bibinfo{person}{Qihao Zhu}, \bibinfo{person}{Shirong Ma}, \bibinfo{person}{Peiyi Wang}, \bibinfo{person}{Xiao Bi}, {et~al\mbox{.}}} \bibinfo{year}{2025}\natexlab{}.
\newblock \showarticletitle{Deepseek-r1: Incentivizing reasoning capability in llms via reinforcement learning}.
\newblock \bibinfo{journal}{\emph{arXiv preprint arXiv:2501.12948}} (\bibinfo{year}{2025}).
\newblock


\bibitem[Ho et~al\mbox{.}(2020)]%
        {xanh2020_2wikimultihop}
\bibfield{author}{\bibinfo{person}{Xanh Ho}, \bibinfo{person}{Anh-Khoa Duong~Nguyen}, \bibinfo{person}{Saku Sugawara}, {and} \bibinfo{person}{Akiko Aizawa}.} \bibinfo{year}{2020}\natexlab{}.
\newblock \showarticletitle{Constructing A Multi-hop {QA} Dataset for Comprehensive Evaluation of Reasoning Steps}. In \bibinfo{booktitle}{\emph{Proceedings of the 28th International Conference on Computational Linguistics}}. \bibinfo{publisher}{International Committee on Computational Linguistics}, \bibinfo{address}{Barcelona, Spain (Online)}, \bibinfo{pages}{6609--6625}.
\newblock
\urldef\tempurl%
\url{https://www.aclweb.org/anthology/2020.coling-main.580}
\showURL{%
\tempurl}


\bibitem[Huang et~al\mbox{.}(2025)]%
        {huang2025rag}
\bibfield{author}{\bibinfo{person}{Jerry Huang}, \bibinfo{person}{Siddarth Madala}, \bibinfo{person}{Risham Sidhu}, \bibinfo{person}{Cheng Niu}, \bibinfo{person}{Julia Hockenmaier}, {and} \bibinfo{person}{Tong Zhang}.} \bibinfo{year}{2025}\natexlab{}.
\newblock \showarticletitle{RAG-RL: Advancing Retrieval-Augmented Generation via RL and Curriculum Learning}.
\newblock \bibinfo{journal}{\emph{arXiv preprint arXiv:2503.12759}} (\bibinfo{year}{2025}).
\newblock


\bibitem[Jin and Yang(2025)]%
        {jin2025scalability}
\bibfield{author}{\bibinfo{person}{Yihong Jin} {and} \bibinfo{person}{Ze Yang}.} \bibinfo{year}{2025}\natexlab{}.
\newblock \showarticletitle{Scalability Optimization in Cloud-Based AI Inference Services: Strategies for Real-Time Load Balancing and Automated Scaling}.
\newblock \bibinfo{journal}{\emph{arXiv preprint arXiv:2504.15296}} (\bibinfo{year}{2025}).
\newblock


\bibitem[Jin et~al\mbox{.}(2024)]%
        {jin2024scam}
\bibfield{author}{\bibinfo{person}{Yihong Jin}, \bibinfo{person}{Ze Yang}, {and} \bibinfo{person}{Xinhe Xu}.} \bibinfo{year}{2024}\natexlab{}.
\newblock \showarticletitle{Scam Detection for Ethereum Smart Contracts: Leveraging Graph Representation Learning for Secure Blockchain}.
\newblock \bibinfo{journal}{\emph{arXiv preprint arXiv:2412.12370}} (\bibinfo{year}{2024}).
\newblock


\bibitem[Kaiser and Weikum(2025)]%
        {kaiser2025preference}
\bibfield{author}{\bibinfo{person}{Magdalena Kaiser} {and} \bibinfo{person}{Gerhard Weikum}.} \bibinfo{year}{2025}\natexlab{}.
\newblock \showarticletitle{Preference-based Learning with Retrieval Augmented Generation for Conversational Question Answering}.
\newblock \bibinfo{journal}{\emph{arXiv preprint arXiv:2503.22303}} (\bibinfo{year}{2025}).
\newblock


\bibitem[Lewis et~al\mbox{.}(2020)]%
        {lewis2020retrieval}
\bibfield{author}{\bibinfo{person}{Patrick Lewis}, \bibinfo{person}{Ethan Perez}, \bibinfo{person}{Aleksandra Piktus}, \bibinfo{person}{Fabio Petroni}, \bibinfo{person}{Vladimir Karpukhin}, \bibinfo{person}{Naman Goyal}, \bibinfo{person}{Heinrich K{\"u}ttler}, \bibinfo{person}{Mike Lewis}, \bibinfo{person}{Wen-tau Yih}, \bibinfo{person}{Tim Rockt{\"a}schel}, {et~al\mbox{.}}} \bibinfo{year}{2020}\natexlab{}.
\newblock \showarticletitle{Retrieval-augmented generation for knowledge-intensive nlp tasks}.
\newblock \bibinfo{journal}{\emph{Advances in neural information processing systems}}  \bibinfo{volume}{33} (\bibinfo{year}{2020}), \bibinfo{pages}{9459--9474}.
\newblock


\bibitem[Li et~al\mbox{.}(2024)]%
        {li2024advances}
\bibfield{author}{\bibinfo{person}{Zilinghan Li}, \bibinfo{person}{Shilan He}, \bibinfo{person}{Ze Yang}, \bibinfo{person}{Minseok Ryu}, \bibinfo{person}{Kibaek Kim}, {and} \bibinfo{person}{Ravi Madduri}.} \bibinfo{year}{2024}\natexlab{}.
\newblock \showarticletitle{Advances in appfl: A comprehensive and extensible federated learning framework}.
\newblock \bibinfo{journal}{\emph{arXiv preprint arXiv:2409.11585}} (\bibinfo{year}{2024}).
\newblock


\bibitem[Liu and Pister(2024)]%
        {liu2024llmeasyquant}
\bibfield{author}{\bibinfo{person}{Dong Liu} {and} \bibinfo{person}{Kaiser Pister}.} \bibinfo{year}{2024}\natexlab{}.
\newblock \showarticletitle{LLMEasyQuant--An Easy to Use Toolkit for LLM Quantization}.
\newblock \bibinfo{journal}{\emph{arXiv preprint arXiv:2406.19657}} (\bibinfo{year}{2024}).
\newblock


\bibitem[Liu et~al\mbox{.}(2024)]%
        {liu2024graphsnapshot}
\bibfield{author}{\bibinfo{person}{Dong Liu}, \bibinfo{person}{Roger Waleffe}, \bibinfo{person}{Meng Jiang}, {and} \bibinfo{person}{Shivaram Venkataraman}.} \bibinfo{year}{2024}\natexlab{}.
\newblock \showarticletitle{Graphsnapshot: Graph machine learning acceleration with fast storage and retrieval}.
\newblock \bibinfo{journal}{\emph{arXiv preprint arXiv:2406.17918}} (\bibinfo{year}{2024}).
\newblock


\bibitem[Liu and Yu(2024)]%
        {liu2024mt2st}
\bibfield{author}{\bibinfo{person}{Dong Liu} {and} \bibinfo{person}{Yanxuan Yu}.} \bibinfo{year}{2024}\natexlab{}.
\newblock \showarticletitle{Mt2st: Adaptive multi-task to single-task learning}.
\newblock \bibinfo{journal}{\emph{arXiv preprint arXiv:2406.18038}} (\bibinfo{year}{2024}).
\newblock


\bibitem[Liu et~al\mbox{.}(2025)]%
        {liu2025roserag}
\bibfield{author}{\bibinfo{person}{Tianci Liu}, \bibinfo{person}{Haoxiang Jiang}, \bibinfo{person}{Tianze Wang}, \bibinfo{person}{Ran Xu}, \bibinfo{person}{Yue Yu}, \bibinfo{person}{Linjun Zhang}, \bibinfo{person}{Tuo Zhao}, {and} \bibinfo{person}{Haoyu Wang}.} \bibinfo{year}{2025}\natexlab{}.
\newblock \showarticletitle{Roserag: Robust retrieval-augmented generation with small-scale llms via margin-aware preference optimization}.
\newblock \bibinfo{journal}{\emph{arXiv preprint arXiv:2502.10993}} (\bibinfo{year}{2025}).
\newblock


\bibitem[Ouyang et~al\mbox{.}(2022)]%
        {ouyang2022training}
\bibfield{author}{\bibinfo{person}{Long Ouyang}, \bibinfo{person}{Jeffrey Wu}, \bibinfo{person}{Xu Jiang}, \bibinfo{person}{Diogo Almeida}, \bibinfo{person}{Carroll Wainwright}, \bibinfo{person}{Pamela Mishkin}, \bibinfo{person}{Chong Zhang}, \bibinfo{person}{Sandhini Agarwal}, \bibinfo{person}{Katarina Slama}, \bibinfo{person}{Alex Ray}, {et~al\mbox{.}}} \bibinfo{year}{2022}\natexlab{}.
\newblock \showarticletitle{Training language models to follow instructions with human feedback}.
\newblock \bibinfo{journal}{\emph{Advances in neural information processing systems}}  \bibinfo{volume}{35} (\bibinfo{year}{2022}), \bibinfo{pages}{27730--27744}.
\newblock


\bibitem[Press et~al\mbox{.}(2022)]%
        {press2022measuring}
\bibfield{author}{\bibinfo{person}{Ofir Press}, \bibinfo{person}{Muru Zhang}, \bibinfo{person}{Sewon Min}, \bibinfo{person}{Ludwig Schmidt}, \bibinfo{person}{Noah~A Smith}, {and} \bibinfo{person}{Mike Lewis}.} \bibinfo{year}{2022}\natexlab{}.
\newblock \showarticletitle{Measuring and narrowing the compositionality gap in language models}.
\newblock \bibinfo{journal}{\emph{arXiv preprint arXiv:2210.03350}} (\bibinfo{year}{2022}).
\newblock


\bibitem[Raffel et~al\mbox{.}(2020)]%
        {raffel2020exploring}
\bibfield{author}{\bibinfo{person}{Colin Raffel}, \bibinfo{person}{Noam Shazeer}, \bibinfo{person}{Adam Roberts}, \bibinfo{person}{Katherine Lee}, \bibinfo{person}{Sharan Narang}, \bibinfo{person}{Michael Matena}, \bibinfo{person}{Yanqi Zhou}, \bibinfo{person}{Wei Li}, {and} \bibinfo{person}{Peter~J Liu}.} \bibinfo{year}{2020}\natexlab{}.
\newblock \showarticletitle{Exploring the limits of transfer learning with a unified text-to-text transformer}.
\newblock \bibinfo{journal}{\emph{Journal of machine learning research}} \bibinfo{volume}{21}, \bibinfo{number}{140} (\bibinfo{year}{2020}), \bibinfo{pages}{1--67}.
\newblock


\bibitem[Robertson et~al\mbox{.}(2009)]%
        {robertson2009probabilistic}
\bibfield{author}{\bibinfo{person}{Stephen Robertson}, \bibinfo{person}{Hugo Zaragoza}, {et~al\mbox{.}}} \bibinfo{year}{2009}\natexlab{}.
\newblock \showarticletitle{The probabilistic relevance framework: BM25 and beyond}.
\newblock \bibinfo{journal}{\emph{Foundations and Trends{\textregistered} in Information Retrieval}} \bibinfo{volume}{3}, \bibinfo{number}{4} (\bibinfo{year}{2009}), \bibinfo{pages}{333--389}.
\newblock


\bibitem[Song et~al\mbox{.}(2025)]%
        {song2025r1}
\bibfield{author}{\bibinfo{person}{Huatong Song}, \bibinfo{person}{Jinhao Jiang}, \bibinfo{person}{Yingqian Min}, \bibinfo{person}{Jie Chen}, \bibinfo{person}{Zhipeng Chen}, \bibinfo{person}{Wayne~Xin Zhao}, \bibinfo{person}{Lei Fang}, {and} \bibinfo{person}{Ji-Rong Wen}.} \bibinfo{year}{2025}\natexlab{}.
\newblock \showarticletitle{R1-Searcher: Incentivizing the Search Capability in LLMs via Reinforcement Learning}.
\newblock \bibinfo{journal}{\emph{arXiv preprint arXiv:2503.05592}} (\bibinfo{year}{2025}).
\newblock


\bibitem[Su et~al\mbox{.}(2022)]%
        {su2022mixed}
\bibfield{author}{\bibinfo{person}{Pei-Chiang Su}, \bibinfo{person}{Shi-Yi Tan}, \bibinfo{person}{Zhenyao Liu}, {and} \bibinfo{person}{Wei-Chang Yeh}.} \bibinfo{year}{2022}\natexlab{}.
\newblock \showarticletitle{A mixed-heuristic quantum-inspired simplified swarm optimization algorithm for scheduling of real-time tasks in the multiprocessor system}.
\newblock \bibinfo{journal}{\emph{Applied Soft Computing}}  \bibinfo{volume}{131} (\bibinfo{year}{2022}), \bibinfo{pages}{109807}.
\newblock


\bibitem[Sun et~al\mbox{.}(2025b)]%
        {SUN2025104147}
\bibfield{author}{\bibinfo{person}{Shiqi Sun}, \bibinfo{person}{Kun Zhang}, \bibinfo{person}{Jingyuan Li}, \bibinfo{person}{Min Yu}, \bibinfo{person}{Kun Hou}, \bibinfo{person}{Yuanzhuo Wang}, {and} \bibinfo{person}{Xueqi Cheng}.} \bibinfo{year}{2025}\natexlab{b}.
\newblock \showarticletitle{Retriever-generator-verification: A novel approach to enhancing factual coherence in open-domain question answering}.
\newblock \bibinfo{journal}{\emph{Information Processing \& Management}} \bibinfo{volume}{62}, \bibinfo{number}{4} (\bibinfo{year}{2025}), \bibinfo{pages}{104147}.
\newblock
\showISSN{0306-4573}
\href{https://doi.org/10.1016/j.ipm.2025.104147}{doi:\nolinkurl{10.1016/j.ipm.2025.104147}}


\bibitem[Sun et~al\mbox{.}(2025a)]%
        {sun2025rearter}
\bibfield{author}{\bibinfo{person}{Zhongxiang Sun}, \bibinfo{person}{Qipeng Wang}, \bibinfo{person}{Weijie Yu}, \bibinfo{person}{Xiaoxue Zang}, \bibinfo{person}{Kai Zheng}, \bibinfo{person}{Jun Xu}, \bibinfo{person}{Xiao Zhang}, \bibinfo{person}{Song Yang}, {and} \bibinfo{person}{Han Li}.} \bibinfo{year}{2025}\natexlab{a}.
\newblock \showarticletitle{ReARTeR: Retrieval-Augmented Reasoning with Trustworthy Process Rewarding}.
\newblock \bibinfo{journal}{\emph{arXiv preprint arXiv:2501.07861}} (\bibinfo{year}{2025}).
\newblock


\bibitem[Touvron et~al\mbox{.}(2023)]%
        {touvron2023llama}
\bibfield{author}{\bibinfo{person}{Hugo Touvron}, \bibinfo{person}{Thibaut Lavril}, \bibinfo{person}{Gautier Izacard}, \bibinfo{person}{Xavier Martinet}, \bibinfo{person}{Marie-Anne Lachaux}, \bibinfo{person}{Timoth{\'e}e Lacroix}, \bibinfo{person}{Baptiste Rozi{\`e}re}, \bibinfo{person}{Naman Goyal}, \bibinfo{person}{Eric Hambro}, \bibinfo{person}{Faisal Azhar}, {et~al\mbox{.}}} \bibinfo{year}{2023}\natexlab{}.
\newblock \showarticletitle{Llama: Open and efficient foundation language models}.
\newblock \bibinfo{journal}{\emph{arXiv preprint arXiv:2302.13971}} (\bibinfo{year}{2023}).
\newblock


\bibitem[Trivedi et~al\mbox{.}(2022a)]%
        {trivedi2022interleaving}
\bibfield{author}{\bibinfo{person}{Harsh Trivedi}, \bibinfo{person}{Niranjan Balasubramanian}, \bibinfo{person}{Tushar Khot}, {and} \bibinfo{person}{Ashish Sabharwal}.} \bibinfo{year}{2022}\natexlab{a}.
\newblock \showarticletitle{Interleaving retrieval with chain-of-thought reasoning for knowledge-intensive multi-step questions}.
\newblock \bibinfo{journal}{\emph{arXiv preprint arXiv:2212.10509}} (\bibinfo{year}{2022}).
\newblock


\bibitem[Trivedi et~al\mbox{.}(2022b)]%
        {trivedi2022musique}
\bibfield{author}{\bibinfo{person}{Harsh Trivedi}, \bibinfo{person}{Niranjan Balasubramanian}, \bibinfo{person}{Tushar Khot}, {and} \bibinfo{person}{Ashish Sabharwal}.} \bibinfo{year}{2022}\natexlab{b}.
\newblock \showarticletitle{MuSiQue: Multihop Questions via Single-hop Question Composition}.
\newblock \bibinfo{journal}{\emph{Transactions of the Association for Computational Linguistics}}  \bibinfo{volume}{10} (\bibinfo{year}{2022}), \bibinfo{pages}{539--554}.
\newblock


\bibitem[Tu et~al\mbox{.}(2025)]%
        {tu2025rbft}
\bibfield{author}{\bibinfo{person}{Yiteng Tu}, \bibinfo{person}{Weihang Su}, \bibinfo{person}{Yujia Zhou}, \bibinfo{person}{Yiqun Liu}, {and} \bibinfo{person}{Qingyao Ai}.} \bibinfo{year}{2025}\natexlab{}.
\newblock \showarticletitle{RbFT: Robust Fine-tuning for Retrieval-Augmented Generation against Retrieval Defects}.
\newblock \bibinfo{journal}{\emph{arXiv preprint arXiv:2501.18365}} (\bibinfo{year}{2025}).
\newblock


\bibitem[Wang et~al\mbox{.}(2020)]%
        {wang2020minilm}
\bibfield{author}{\bibinfo{person}{Wenhui Wang}, \bibinfo{person}{Furu Wei}, \bibinfo{person}{Li Dong}, \bibinfo{person}{Hangbo Bao}, \bibinfo{person}{Nan Yang}, {and} \bibinfo{person}{Ming Zhou}.} \bibinfo{year}{2020}\natexlab{}.
\newblock \showarticletitle{Minilm: Deep self-attention distillation for task-agnostic compression of pre-trained transformers}.
\newblock \bibinfo{journal}{\emph{Advances in neural information processing systems}}  \bibinfo{volume}{33} (\bibinfo{year}{2020}), \bibinfo{pages}{5776--5788}.
\newblock


\bibitem[Wang et~al\mbox{.}(2024b)]%
        {wang2024maferw}
\bibfield{author}{\bibinfo{person}{Yujing Wang}, \bibinfo{person}{Hainan Zhang}, \bibinfo{person}{Liang Pang}, \bibinfo{person}{Binghui Guo}, \bibinfo{person}{Hongwei Zheng}, {and} \bibinfo{person}{Zhiming Zheng}.} \bibinfo{year}{2024}\natexlab{b}.
\newblock \showarticletitle{MaFeRw: Query rewriting with multi-aspect feedbacks for retrieval-augmented large language models}.
\newblock \bibinfo{journal}{\emph{arXiv preprint arXiv:2408.17072}} (\bibinfo{year}{2024}).
\newblock


\bibitem[Wang et~al\mbox{.}(2024a)]%
        {wang2024speculative}
\bibfield{author}{\bibinfo{person}{Zilong Wang}, \bibinfo{person}{Zifeng Wang}, \bibinfo{person}{Long Le}, \bibinfo{person}{Huaixiu~Steven Zheng}, \bibinfo{person}{Swaroop Mishra}, \bibinfo{person}{Vincent Perot}, \bibinfo{person}{Yuwei Zhang}, \bibinfo{person}{Anush Mattapalli}, \bibinfo{person}{Ankur Taly}, \bibinfo{person}{Jingbo Shang}, {et~al\mbox{.}}} \bibinfo{year}{2024}\natexlab{a}.
\newblock \showarticletitle{Speculative rag: Enhancing retrieval augmented generation through drafting}.
\newblock \bibinfo{journal}{\emph{arXiv preprint arXiv:2407.08223}} (\bibinfo{year}{2024}).
\newblock


\bibitem[Wei et~al\mbox{.}(2024)]%
        {wei2024instructrag}
\bibfield{author}{\bibinfo{person}{Zhepei Wei}, \bibinfo{person}{Wei-Lin Chen}, {and} \bibinfo{person}{Yu Meng}.} \bibinfo{year}{2024}\natexlab{}.
\newblock \showarticletitle{InstructRAG: Instructing Retrieval-Augmented Generation via Self-Synthesized Rationales}.
\newblock \bibinfo{journal}{\emph{arXiv preprint arXiv:2406.13629}} (\bibinfo{year}{2024}).
\newblock


\bibitem[Xiang et~al\mbox{.}(2024)]%
        {xiang2024neural}
\bibfield{author}{\bibinfo{person}{Ao Xiang}, \bibinfo{person}{Bingjie Huang}, \bibinfo{person}{Xinyu Guo}, \bibinfo{person}{Haowei Yang}, {and} \bibinfo{person}{Tianyao Zheng}.} \bibinfo{year}{2024}\natexlab{}.
\newblock \showarticletitle{A neural matrix decomposition recommender system model based on the multimodal large language model}. In \bibinfo{booktitle}{\emph{Proceedings of the 2024 7th International Conference on Machine Learning and Machine Intelligence (MLMI)}}. \bibinfo{pages}{146--150}.
\newblock


\bibitem[Xiong et~al\mbox{.}(2025)]%
        {xiong2025rag}
\bibfield{author}{\bibinfo{person}{Guangzhi Xiong}, \bibinfo{person}{Qiao Jin}, \bibinfo{person}{Xiao Wang}, \bibinfo{person}{Yin Fang}, \bibinfo{person}{Haolin Liu}, \bibinfo{person}{Yifan Yang}, \bibinfo{person}{Fangyuan Chen}, \bibinfo{person}{Zhixing Song}, \bibinfo{person}{Dengyu Wang}, \bibinfo{person}{Minjia Zhang}, {et~al\mbox{.}}} \bibinfo{year}{2025}\natexlab{}.
\newblock \showarticletitle{Rag-gym: Optimizing reasoning and search agents with process supervision}.
\newblock \bibinfo{journal}{\emph{arXiv preprint arXiv:2502.13957}} (\bibinfo{year}{2025}).
\newblock


\bibitem[Yang et~al\mbox{.}(2025b)]%
        {yang2025qwen3}
\bibfield{author}{\bibinfo{person}{An Yang}, \bibinfo{person}{Anfeng Li}, \bibinfo{person}{Baosong Yang}, \bibinfo{person}{Beichen Zhang}, \bibinfo{person}{Binyuan Hui}, \bibinfo{person}{Bo Zheng}, \bibinfo{person}{Bowen Yu}, \bibinfo{person}{Chang Gao}, \bibinfo{person}{Chengen Huang}, \bibinfo{person}{Chenxu Lv}, {et~al\mbox{.}}} \bibinfo{year}{2025}\natexlab{b}.
\newblock \showarticletitle{Qwen3 Technical Report}.
\newblock \bibinfo{journal}{\emph{arXiv preprint arXiv:2505.09388}} (\bibinfo{year}{2025}).
\newblock


\bibitem[Yang et~al\mbox{.}(2025a)]%
        {yang2025data}
\bibfield{author}{\bibinfo{person}{Qikai Yang}, \bibinfo{person}{Cheng Ji}, \bibinfo{person}{Huaiying Luo}, \bibinfo{person}{Panfeng Li}, {and} \bibinfo{person}{Zhicheng Ding}.} \bibinfo{year}{2025}\natexlab{a}.
\newblock \showarticletitle{Data Augmentation Through Random Style Replacement}.
\newblock \bibinfo{journal}{\emph{arXiv preprint arXiv:2504.10563}} (\bibinfo{year}{2025}).
\newblock


\bibitem[Yang et~al\mbox{.}(2018)]%
        {yang2018hotpotqa}
\bibfield{author}{\bibinfo{person}{Zhilin Yang}, \bibinfo{person}{Peng Qi}, \bibinfo{person}{Saizheng Zhang}, \bibinfo{person}{Yoshua Bengio}, \bibinfo{person}{William~W Cohen}, \bibinfo{person}{Ruslan Salakhutdinov}, {and} \bibinfo{person}{Christopher~D Manning}.} \bibinfo{year}{2018}\natexlab{}.
\newblock \showarticletitle{HotpotQA: A dataset for diverse, explainable multi-hop question answering}.
\newblock \bibinfo{journal}{\emph{arXiv preprint arXiv:1809.09600}} (\bibinfo{year}{2018}).
\newblock


\bibitem[Yao et~al\mbox{.}(2023)]%
        {yao2023react}
\bibfield{author}{\bibinfo{person}{Shunyu Yao}, \bibinfo{person}{Jeffrey Zhao}, \bibinfo{person}{Dian Yu}, \bibinfo{person}{Nan Du}, \bibinfo{person}{Izhak Shafran}, \bibinfo{person}{Karthik Narasimhan}, {and} \bibinfo{person}{Yuan Cao}.} \bibinfo{year}{2023}\natexlab{}.
\newblock \showarticletitle{React: Synergizing reasoning and acting in language models}. In \bibinfo{booktitle}{\emph{International Conference on Learning Representations (ICLR)}}.
\newblock


\bibitem[Zhang et~al\mbox{.}(2025b)]%
        {zhang2025rag}
\bibfield{author}{\bibinfo{person}{Hanning Zhang}, \bibinfo{person}{Juntong Song}, \bibinfo{person}{Juno Zhu}, \bibinfo{person}{Yuanhao Wu}, \bibinfo{person}{Tong Zhang}, {and} \bibinfo{person}{Cheng Niu}.} \bibinfo{year}{2025}\natexlab{b}.
\newblock \showarticletitle{RAG-Reward: Optimizing RAG with Reward Modeling and RLHF}.
\newblock \bibinfo{journal}{\emph{arXiv preprint arXiv:2501.13264}} (\bibinfo{year}{2025}).
\newblock


\bibitem[Zhang et~al\mbox{.}(2024a)]%
        {zhang2024sirerag}
\bibfield{author}{\bibinfo{person}{Nan Zhang}, \bibinfo{person}{Prafulla~Kumar Choubey}, \bibinfo{person}{Alexander Fabbri}, \bibinfo{person}{Gabriel Bernadett-Shapiro}, \bibinfo{person}{Rui Zhang}, \bibinfo{person}{Prasenjit Mitra}, \bibinfo{person}{Caiming Xiong}, {and} \bibinfo{person}{Chien-Sheng Wu}.} \bibinfo{year}{2024}\natexlab{a}.
\newblock \showarticletitle{SiReRAG: Indexing Similar and Related Information for Multihop Reasoning}.
\newblock \bibinfo{journal}{\emph{arXiv preprint arXiv:2412.06206}} (\bibinfo{year}{2024}).
\newblock


\bibitem[Zhang et~al\mbox{.}(2025a)]%
        {zhang2025automated}
\bibfield{author}{\bibinfo{person}{Zheyu Zhang}, \bibinfo{person}{Yutong Luo}, \bibinfo{person}{Yongzhou Chen}, \bibinfo{person}{Haopeng Zhao}, \bibinfo{person}{Zhichao Ma}, {and} \bibinfo{person}{Hao Liu}.} \bibinfo{year}{2025}\natexlab{a}.
\newblock \showarticletitle{Automated Parking Trajectory Generation Using Deep Reinforcement Learning}.
\newblock \bibinfo{journal}{\emph{arXiv preprint arXiv:2504.21071}} (\bibinfo{year}{2025}).
\newblock


\bibitem[Zhang et~al\mbox{.}(2024b)]%
        {zhang2024trustworthy}
\bibfield{author}{\bibinfo{person}{Zongmeng Zhang}, \bibinfo{person}{Yufeng Shi}, \bibinfo{person}{Jinhua Zhu}, \bibinfo{person}{Wengang Zhou}, \bibinfo{person}{Xiang Qi}, \bibinfo{person}{Peng Zhang}, {and} \bibinfo{person}{Houqiang Li}.} \bibinfo{year}{2024}\natexlab{b}.
\newblock \showarticletitle{Trustworthy alignment of retrieval-augmented large language models via reinforcement learning}.
\newblock \bibinfo{journal}{\emph{arXiv preprint arXiv:2410.16843}} (\bibinfo{year}{2024}).
\newblock


\bibitem[Zhao and Chen(2025)]%
        {zhao2025llm}
\bibfield{author}{\bibinfo{person}{Chuqing Zhao} {and} \bibinfo{person}{Yisong Chen}.} \bibinfo{year}{2025}\natexlab{}.
\newblock \showarticletitle{LLM-powered Topic Modeling for Discovering Public Mental Health Trends in Social Media}.
\newblock  (\bibinfo{year}{2025}).
\newblock


\bibitem[Zhao et~al\mbox{.}(2025)]%
        {zhao2025optimizedpathplanninglogistics}
\bibfield{author}{\bibinfo{person}{Haopeng Zhao}, \bibinfo{person}{Zhichao Ma}, \bibinfo{person}{Lipeng Liu}, \bibinfo{person}{Yang Wang}, \bibinfo{person}{Zheyu Zhang}, {and} \bibinfo{person}{Hao Liu}.} \bibinfo{year}{2025}\natexlab{}.
\newblock \showarticletitle{Optimized Path Planning for Logistics Robots Using Ant Colony Algorithm under Multiple Constraints}.
\newblock \bibinfo{journal}{\emph{arXiv preprint arXiv:2504.05339}} (\bibinfo{year}{2025}).
\newblock


\bibitem[Zhu et~al\mbox{.}(2025)]%
        {zhu2025mitigating}
\bibfield{author}{\bibinfo{person}{Rongzhi Zhu}, \bibinfo{person}{Xiangyu Liu}, \bibinfo{person}{Zequn Sun}, \bibinfo{person}{Yiwei Wang}, {and} \bibinfo{person}{Wei Hu}.} \bibinfo{year}{2025}\natexlab{}.
\newblock \showarticletitle{Mitigating Lost-in-Retrieval Problems in Retrieval Augmented Multi-Hop Question Answering}.
\newblock \bibinfo{journal}{\emph{arXiv preprint arXiv:2502.14245}} (\bibinfo{year}{2025}).
\newblock


\end{thebibliography}

\appendix

\end{document}